\documentclass[twocolumn]{article}
\usepackage{epsfig}
\usepackage{booktabs}
\usepackage{amssymb,amsmath,amsthm,multirow}
\usepackage{amsfonts,graphicx,subfigure}
\usepackage{float}
\usepackage{caption}
\usepackage{subcaption}
\usepackage{algorithm}
\usepackage{algpseudocode}
\usepackage{xcolor} 
\usepackage[table]{xcolor} 
\newtheorem{theorem}{Theorem}

\newtheorem{lemma}{Lemma}
\newtheorem{proposition}{Proposition}

\begin{document}
\title{Improving clustering quality evaluation in noisy Gaussian mixtures}
\author{Renato Cordeiro de Amorim\thanks{School of Computer Science and
Electronic Engineering, University of Essex, Wivenhoe, UK. r.amorim@essex.ac.uk} \and Vladimir Makarenkov\thanks{D\'epartement d'informatique, Universit\'e du Québec \`a Montr\'eal, C.P. 8888 succ. Centre-Ville, Montreal (QC) H3C 3P8 Canada.} \thanks{Mila - Quebec AI Institute, Montreal, QC, Canada.}} 
\date{}
\maketitle

\begin{abstract}
Clustering is a well-established technique in machine learning and data analysis, widely used across various domains. Cluster validity indices, such as the Average Silhouette Width, Calinski-Harabasz, and Davies-Bouldin indices, play a crucial role in assessing clustering quality when external ground truth labels are unavailable. However, these measures can be affected by different degrees of feature relevance, potentially leading to unreliable evaluations in high-dimensional or noisy data sets.

We introduce a theoretically grounded Feature Importance Rescaling (FIR) method that enhances the quality of clustering validation by adjusting feature contributions based on their dispersion. It attenuates noise features, clarifies clustering compactness and separation, and thereby aligns clustering validation more closely with the ground truth. Through extensive experiments on synthetic data sets under different configurations and a case study on real-world data, we demonstrate that FIR consistently improves the correlation between the values of cluster validity indices and the ground truth, particularly in settings with noisy or irrelevant features.

The results show that FIR increases the robustness of clustering evaluation, reduces variability in performance across different data sets, and remains effective even when clusters exhibit significant overlap. These findings highlight the potential of FIR as a valuable enhancement of clustering validation, making it a practical tool for unsupervised learning tasks where labelled data is unavailable.
\end{abstract}

\section{Introduction}
Clustering is a fundamental technique in machine learning and data analysis, which is central to many exploratory methods. It aims at forming homogeneous data groups (i.e., clusters), according to a selected similarity measure, without requiring labels to learn from. Clustering algorithms have been successfully applied to solve many practical problems from various application fields, including data mining, community detection, computer vision, and natural language processing \cite{mirkin2022community,mittal2022comprehensive,ikotun2023k,zampieri2014between}.

There are different approaches to clustering that algorithms may employ. For instance, partitional clustering algorithms generate a clustering with non-overlapping clusters that collectively cover all data points (i.e. a partition of the data). Hierarchical algorithms iteratively merge (agglomerative) or split (divisive) clusters, producing a tree-like structure that can be visualised with a dendrogram representing both the clustering and the relationships between clusters. In this, a data point may belong to more than one cluster as long as these memberships happen at different levels of the hierarchy. Fuzzy clustering algorithms allow each data point to belong to more than one cluster, with degrees of membership usually adding to one. For more details on these and other approaches, we direct interested readers to the literature (see, for instance, \cite{ran2023comprehensive,oyewole2023data} and references therein).

Here, we focus on the internal evaluation of clusterings that are non-overlapping partitions of a data set (such partitions are sometimes called a crisp clustering). Internal evaluation assesses clustering quality without relying on external factors, such as ground truth labels. Instead, it considers only the intrinsic properties of the data and the resulting clustering. Key aspects include within-cluster cohesion (compactness of clusters) and between-cluster separation (degree of distinction between clusters). This aligns well with real-world clustering applications, where labels are typically unavailable. Internal evaluation has been extensively studied in the literature \cite{arbelaitz2013extensive,todeschini2024extended,rykov2024inertia}.

The contribution of this paper is a theoretically sound method for enhancing internal evaluation measures by accounting for feature relevance. Our approach, called Feature Importance Rescaling (FIR), recognises that different features may have different degrees of relevance, and applies these to rescale a data set. We demonstrate that our rescaling improves the correlation between four popular internal evaluation measures and ground truth labels.

\section{Related work}
\label{sec:related_work}
The $k$-means algorithm \cite{macqueen1967some} is arguably the most popular clustering algorithm there is \cite{jaeger2023cluster,jain2010data}. Given a data set $X=\{x_1, \ldots, x_n\}$, where each $x_i \in X$ is described over $m$ features, $k$-means produces a clustering $C=\{C_1, \ldots, C_k\}$ by iteratively minimising the Within-Cluster Sum of Squares (WCSS),
\begin{equation}
\label{eq:kmeans}
    WCSS = \sum_{l=1}^k \sum_{x_i \in C_l} d(x_i, z_l),
\end{equation}
where $z_l$ is the centroid of cluster $C_l \in C$, and $d(x_i,z_l)$ is the Euclidean distance between $x_i$ and $z_l$. Algorithm \ref{alg:kmeans} details the steps.

\begin{algorithm}
    \caption{$k$-means}
    \label{alg:kmeans}
    \begin{algorithmic}[1]
        \Require Data set $X$, number of clusters $k$.
        \Ensure Clustering $C = \{C_1, \dots, C_k\}$ and centroids $Z = \{z_1, \ldots, z_k\}$
        \State Select $k$ data points from $X$ uniformly at random, and copy their values into $z_1, \ldots, z_k$.
        \Repeat
            \State Assign each $x_i \in X$ to the cluster of its nearest centroid. That is,
            \[
            C_l \gets \{x_i \in X \mid l=\arg\min_{t} d(x_i, z_t)\}.
            \]
            \State Update each $z_l \in Z$ to the component-wise mean of $x_i \in C_l$.
        \Until{centroids do not change.}
        \State \Return Clustering $C$ and centroids $Z$.
    \end{algorithmic}
\end{algorithm}
The clustering $C$ is a partition of $X$. Hence, $X=\bigcup_{l=1}^k C_l$, and $C_l \cap C_t = \emptyset$ for all $C_l,C_t \in C$ with $l\neq t$. $K$-means initialises its centroids randomly and makes locally optimal choices at each iteration. As a result, it is a non-deterministic algorithm, and its final clustering heavily depends on the quality of the initial centroids. Considerable research has focused on identifying better initial centroids (see, for instance, \cite{harris2022extensive,franti2019much}, and references therein), with $k$-means++ \cite{arthur2007k} being the most widely adopted method. The latter employs a probabilistic centroid selection mechanism favouring distant points as initial centroids (see Algorithm \ref{alg:kmeanspp}). In fact, many software packages, including scikit-learn, MATLAB, and R, use $k$-means++ as the default initialisation for $k$-means.
\begin{algorithm}
    \caption{$k$-means++}
    \label{alg:kmeanspp}
    \begin{algorithmic}[1]
        \Require Data set $X = \{x_1, \ldots, x_n\}$, number of clusters $k$.
        \Ensure Clustering $C = \{C_1, \dots, C_k\}$ and centroids $Z = \{z_1, \ldots, z_k\}$.
        \State Select the first centroid $z_1$ uniformly at random from $X$.
        \For{$l = 2$ to $k$}
            \State Compute the distance of each $x_i \in X$ to its closest currently chosen centroid:
            \[
            D(x_i) = \min_{1 \leq t < l} d(x_i, z_t).
            \]
            \State Randomly select the next centroid $z_l$ from $X$ by the probability distribution:
            \[
            P(x_i) = \frac{D(x_i)}{\sum_{x_j \in X} D(x_j)}.
            \]
        \EndFor
        \State Run $k$-means (Algorithm \ref{alg:kmeans}) using $Z$ as initial centroids.
        \State \Return Clustering $C$ and final centroids $Z$.
    \end{algorithmic}
\end{algorithm}

Despite its effectiveness, $k$-means++ is not without limitations. Due to its inherent randomness in centroid selection, it is typically executed multiple times, potentially producing different clustering outcomes. This raises a fundamental question addressed in this paper: given multiple clusterings, how should the most suitable one be selected? The literature suggests various approaches. If the number of clusters, $k$, is fixed one can select the clustering minimising the WCSS in (\ref{eq:kmeans}) as the final clustering. Another approach, particularly useful if $k$ is unknown, is to employ a cluster validity index to evaluate the quality of each clustering.

\subsection{Cluster validity indices}

Cluster validity indices are measures used to evaluate the quality of clusterings, examining both the cluster assignments and the underlying data structure. A large number of such indices have been proposed in the literature, including many recent extensions and specialised criteria \cite{todeschini2024extended}. An exhaustive comparison with all available indices would be impractical. In this work we therefore focus on a small set of widely used classical indices. In particular, the Silhouette width, Calinski-Harabasz, and Davies-Bouldin indices consistently exhibit strong performance across diverse applications \cite{arbelaitz2013extensive}, making them representative benchmarks for assessing the effect of FIR. Hence, this section focuses on these three measures.

The Silhouette width \cite{rousseeuw1987silhouettes} of a data point $x_i$, $s(x_i)$, is given by
\begin{equation}
    s(x_i) = \frac{b(x_i) - a(x_i)}{\text{max}\{a(x_i), b(x_i)\}},
\end{equation}
where $a(x_i)$ is the average distance between $x_i \in C_l$ and all $x_j \in C_l$ with $i \neq j$. That is,
\begin{equation*}
    a(x_i) = \frac{1}{|C_l|-1}\sum_{x_j \in C_l, i\neq j}d(x_i, x_j).
\end{equation*}

The value $b(x_i)$ represents the lowest average distance between $x_i \in C_l$ and all points in any other cluster. Formally,
\begin{equation*}
    b(x_i) = \text{min}_{t\neq l} \frac{1}{|C_t|}\sum_{x_j \in C_t}d(x_i, x_j).
\end{equation*}

The coefficient $s(x_i)$ defines the value of the Silhouette width for a particular point $x_i$. In order to determine the value of this index for the whole clustering, we need to calculate the Average Silhouette Width (ASW),
\begin{equation}
    ASW = \frac{1}{n}\sum_{i=1}^n s(x_i).
\end{equation}

The Silhouette width has some interesting properties. For each data point \(x_i \in X\), its Silhouette value \(s(x_i)\) is bounded between \(-1\) and \(1\). A value near \(1\) indicates that \(x_i\) is well-matched to its assigned cluster and is distinctly separated from other clusters, whereas a value near \(-1\) suggests a potential misclassification. Although in $k$-means we employ the Euclidean distance, the Silhouette width is distance metric agnostic. This is a particularly useful property when assessing a clustering formed under a metric other than the Euclidean distance.

The Calinski-Harabasz index (CH) \cite{calinski1974dendrite} is another popular cluster validity index. It quantifies the ratio between between-cluster dispersion and within-cluster dispersion. First, the Between-Cluster Sum of Squares (BCSS) is given by

\begin{equation*}
    BCSS(C)= \sum_{l=1}^k|C_l| \cdot d(z_l, c),
\end{equation*}
where $c$ is the component-wise mean calculated over all $x_i \in X$, and $z_l$ is the centroid of cluster $C_l$. Second, the Within-Cluster Sum of Squares (WCSS) is calculated using Equation (\ref{eq:kmeans}). The value of this index for a given clustering $C$ is defined as

\begin{equation*}
    CH = \frac{BCSS/(k-1)}{WCSS/(n-k)}.
\end{equation*}

The value of CH is high when clusters are both well separated (large BCSS) and compact (small WCSS), indicating a more distinct clustering structure. Note that computing this index involves primarily calculating centroids and the associated sums of squares, making it efficient to compute even for large data sets.

The Davies-Bouldin index (DB) \cite{davies1979cluster} evaluates clustering quality by quantifying the trade-off between within-cluster compactness and between-cluster separation. For each cluster $C_l\in C$, we first compute its within-cluster scatter $S_l$, defined as the average distance between points in \( C_l \) from the centroid \( z_l \). That is,
\begin{equation*}
        S_l = \frac{1}{|C_l|} \sum_{x_i \in C_l} d(x_i, z_l).
\end{equation*}

Then, for every pair of distinct clusters \( C_l \) and \( C_t \) (with centroids \( z_c \) and \( z_t \), respectively), we calculate the similarity measure:
\begin{equation*}
        R_{lt} = \frac{S_l + S_t}{d(z_l, z_t)}.
\end{equation*}

For each cluster \( C_l \in C\), we determine the worst-case (i.e. maximum) ratio with respect to all other clusters:
\begin{equation*}
    R_l = \max_{t \neq l} R_{lt}.
\end{equation*}
Finally, the Davies-Bouldin index for the clustering \( C \) is the average of these worst-case ratios:
\begin{equation*}
        DB = \frac{1}{k} \sum_{l=1}^k R_l.
    \end{equation*}

Lower values of \(DB\) indicate better clustering, as they reflect clusters that are both compact (low \( S_l \)) and well separated (high \( d(z_l, z_t) \)).

\section{Feature importance rescaling}
\label{sec:new_method}

In this section, we introduce our Feature Importance Rescaling (FIR) method. This data rescaling method was designed to enhance the evaluation of clustering quality performed by the measures discussed in Section \ref{sec:related_work}, as well as WCSS (\ref{eq:kmeans}). Our approach achieves this by quantifying the relevance of features and by using this information to rescale the data set accordingly. This method is particularly suited for partitional clustering algorithms, such as $k$-means++, which assume the data follows a Gaussian distribution. That is, data points are concentrated around the cluster centroid.

The $k$-means++ algorithm iteratively minimises the within-cluster sum of squares, given by Equation (\ref{eq:kmeans}). If we are to apply a rescaling factor $\alpha_{v}$ to each feature $v$, the objective function becomes

\begin{align}
\label{eq:weighted_dispersion}
\begin{split}
    WCSS_w &= \sum_{l=1}^k \sum_{x_i \in C_l} \sum_{v=1}^m (\alpha_{v}x_{iv} - \alpha_{v}z_{lv})^2 \\
    &= \sum_{l=1}^k \sum_{x_i \in C_l} \sum_{v=1}^m \alpha_{v}^2 (x_{iv} - z_{lv})^2 \\
    &= \sum_{v=1}^m \alpha_{v}^2 \sum_{l=1}^k \sum_{x_i \in C_l} (x_{iv} - z_{lv})^2 \\
    &= \sum_{v=1}^m \alpha_{v}^2 D_{v},
\end{split}
\end{align}

where $D_{v}$ is the dispersion of feature $v$,

\begin{equation}
    \label{eq:dispersion}
     D_{v} = \sum_{l=1}^k\sum_{x_i \in C_l} (x_{iv}- z_{lv})^2 + \epsilon,
\end{equation}
and $\epsilon>0$ is a small numerical floor ensuring $D_v>0$. In practice we set $\epsilon = 10^{-3}$ after range normalisation (see Eq.~\ref{eq:range_normalisation}). Features that are globally constant are removed during preprocessing, so a feature with $D_v\approx 0$ indicates it is (near) constant within clusters but distinct across clusters.  Our method then (appropriately) produces a high weight while normalisation $\sum_v \alpha_v=1$ prevents unbounded dominance.

Minimising $D_{v}$ aligns well with the optimisation objective of partitional clustering algorithms, which seek to reduce within-cluster variance while maintaining between-cluster separation. That is, we aim to rescale the data favouring features with a low dispersion. To determine the optimal feature rescaling factors $\alpha_{v}$, we devise a Lagrangian function with a constraint ensuring that the sum of the rescaling factors equals one,

\begin{equation}
    \mathcal{L} = \sum_{v=1}^m \alpha_{v}^2 D_{v} + \lambda\left(\sum_{v=1}^m \alpha_{v} - 1 \right).
\end{equation}

Taking partial derivatives with respect to $\alpha_{v}$ and the Lagrange multiplier $\lambda$, we obtain

\begin{align}
    \frac{\partial \mathcal{L}}{\partial \alpha_{v}} &= \alpha_{v} D_{v} + \lambda = 0, \label{eq:1st_partial} \\
    \frac{\partial \mathcal{L}}{\partial \lambda} &= \sum_{v=1}^m \alpha_{v} - 1 = 0. \label{eq:2nd_partial}
\end{align}

Solving Equation (\ref{eq:1st_partial}) for $\alpha_{v}$, we get

\begin{equation}
    \alpha_{v} = \frac{-\lambda}{D_{v}}.
\end{equation}

Substituting this into Equation (\ref{eq:2nd_partial})

\begin{equation}
    \sum_{j=1}^m \frac{-\lambda}{D_{j}} = 1 \quad \iff \quad -\lambda = \frac{1}{\sum_{j=1}^m D_{j}}.
\end{equation}

Thus, the optimal rescaling factor for a feature $v$ is given by

\begin{equation}
    \label{eq:rescaling_factor}
    \alpha_{v} = \frac{1}{\sum_{j=1}^m \frac{D_{v}}{D_{j}}}.
\end{equation}

A feature $v$ is considered more relevant to a clustering solution when it contributes significantly to defining cluster structure. Since clustering methods attempt to minimise within-cluster variance, a natural way to quantify relevance is to assume that features with lower dispersion (measured with Equation \ref{eq:dispersion}) should be given a higher rescaling factor, leading to higher importance. Hence, our rescaling method dynamically adapts feature importance, ensuring that cluster quality evaluation measures operate in a space where informative features are emphasized while noisy or less relevant features are attenuated. Empirical results suggest that applying our method twice to a data set often improves performance slightly. Hence, this is how we formally describe the method in Algorithm \ref{alg:new_method}.

\begin{algorithm}
    \caption{Feature Importance Rescaling}
    \label{alg:new_method}
    \begin{algorithmic}[1]
        \Require Dataset $X$, clustering $C=\{C_1, \ldots, C_k\}$, number of iterations $iter$ (we suggest 2).
        \Ensure A rescaled data set $X^\prime$.
        \State Set $m$ to be the number of features in $X$.
        \For{$i=1$ to $iter$}
            \State Compute each centroid $z_l \in \{z_1, \ldots, z_k\}$ as the component-wise mean of $x_i \in C_l$.
            \For{$v = 1$ to $m$}
                \State Compute $\alpha_v$ using Equation (\ref{eq:rescaling_factor}).
                \State Set $X_v^\prime=\alpha_v\cdot X_v$, where $X_v$ represents feature $v$ over all points in $X$.
            \EndFor
        \EndFor
        \State \Return $X^\prime$.
    \end{algorithmic}
\end{algorithm}

It is important to note that FIR is specifically designed for partitional clustering algorithms that minimise within-cluster variance, most notably $k$-means and its $k$-means++ initialisation. These are by far the most widely used clustering algorithms in both research and applied machine learning, which makes them a natural focus for our study. In addition, FIR evaluates feature relevance through the marginal within-cluster dispersion \(D_v\), and therefore does not explicitly model correlations or covariance structure between features. In high-dimensional settings where informative features are strongly correlated, marginal dispersion alone may not fully capture their joint contribution to cluster structure. These design choices keep FIR simple, interpretable, and computationally lightweight, while remaining compatible with common clustering objectives such as that of $k$-means. Extensions incorporating richer dependency information (such as covariance-aware formulations based on cluster-wise covariance matrices or Mahalanobis-type measures) represent interesting directions for future work.

\subsection{Relation to feature selection methods}

Before turning to the theoretical properties of our method, it is important to distinguish FIR from other approaches that also aim to address feature relevance. Classical examples include ReliefF~\cite{kira1992practical,kononenko1994estimating}, mRMR (minimum Redundancy Maximum Relevance)~\cite{peng2005feature}, Laplacian Score~\cite{he2005laplacian}, and Fisher Score~\cite{gu2012generalized}. These methods have been widely applied in high-dimensional learning problems ranging from bioinformatics to text mining, and remain highly influential in the literature.

A key difference lies in the type of output they produce. ReliefF, mRMR, Laplacian Score, and Fisher Score are primarily feature selection methods. They assign each feature a relevance score and then construct a reduced data set by retaining only those features deemed informative. The effect is a change in the representation itself, as irrelevant features are removed entirely.

By contrast, our objective is not to eliminate features. FIR preserves every feature in the representation, but its contribution is modulated by a continuous factor derived from its dispersion. This distinction is crucial in the context of clustering evaluation. Cluster validity indices such as ASW, CH, DB, and WCSS are defined on the full feature space. If features are discarded, the definition of these indices changes because distances, dispersions, and separations are computed in a different space. Our approach avoids this problem: FIR leaves the feature space intact, ensuring that the indices remain well defined, while attenuating the influence of noisy or irrelevant dimensions.

Another distinction is methodological. Feature selection algorithms typically rely on criteria such as mutual information, statistical dependence, or manifold structure, and in many cases assume access to class labels. In contrast, FIR is unsupervised and parameter-free. It does not require labels, external supervision, or additional hyperparameters. Its role is specifically to improve the alignment between internal cluster validity indices and external clustering quality, a setting not addressed by the traditional feature selection methods mentioned above.

In this sense, FIR and feature selection are complementary rather than competing approaches. Feature selection methods reduce the dimensionality of the data prior to clustering, whereas FIR adjusts the contribution of all features when evaluating clustering quality. Both address the challenge of high-dimensional and noisy data, but they do so at different stages of the analysis pipeline and under different objectives.

\subsection{Theoretical Properties}

In this section, we establish several theoretical properties of the FIR method. We begin by proving that FIR is a computationally free enhancement to $k$-means and $k$-means++.

\begin{theorem}
\label{thm:free}
    FIR is an asymptotically free enhancement to $k$-means and $k$-means++.
\end{theorem}
\begin{proof}
First, we quantify the cost of a single FIR iteration. Given that each data point contributes once to exactly one centroid update, we have that computing $k$ centroids takes $O(nm)$ time. Evaluating the dispersions in Equation (\ref{eq:dispersion}) for each feature is another pass over the data and costs $O(nm)$. Given the precomputed sum $S=\sum_{j=1}^m \frac{1}{D_j}$, the weights $\alpha_v=\frac{1}{D_v S}$ for all $v$ are obtained in $O(m)$. 
Rescaling feature $v$ across all $n$ points costs $O(n)$, hence $O(nm)$ for all features. 
Thus, one FIR iteration is $O(nm)$, and since the number of FIR iterations is a fixed constant (e.g., $2$), FIR in total is $O(nm)$.

In contrast, each iteration of $k$-means or $k$-means++ assigns all $n$ points to their nearest of $k$ centroids in $m$ dimensions, taking $O(nkm)$ time; with $\tau$ iterations until convergence, the clustering phase costs $O(\tau n k m)$. 
Therefore, the combined cost of FIR and either $k$-means or $k$-means++ is
\[
O(nm) + O(\tau n k m) = O(\tau n k m).
\]
That is, adding FIR does not change the asymptotic time complexity of the overall procedure. Hence, FIR is computationally free in this scenario.
\end{proof}

We now show that the FIR objective function, \(WCSS_w\), is strictly convex under usual conditions, and that the optimisation problem admits a unique solution.
\begin{theorem}
    \(WCSS_w\) is convex, and FIR provides a unique solution for any data set containing non-trivial features.
\end{theorem}
\begin{proof}
    We define a feature $v$ as trivial if $D_v =0$, since such a feature should be removed during data pre-processing. The objective function is given by \(WCSS_w = \sum_{v=1}^m \alpha_v^2D_v\), with second partial derivative
    \begin{align*}
        \frac{\partial^2 WCSS_w}{\partial^2 \alpha^2_v}=2D_v.
    \end{align*}
    Given $D_v> 0$ for all $v$, the second derivative is positive. Hence, the objective is convex. The Hessian of \(WCSS_w\) is the diagonal matrix
    \[
    \nabla^2 WCSS_w = \text{diag}(2D_1, 2D_2, \ldots, 2D_m),
    \]
    which is positive definite. Hence, \(WCSS_w\) is strictly convex on \(\mathbb{R}^m\). 
    
    The constraint \(\sum_{v=1}^m \alpha_v=1\) defines a non-empty affine subspace, which is convex. The minimisation of a strictly convex function over a convex set has a unique global minimiser.    
\end{proof}

Next, we provide a fundamental theoretical interpretation of the FIR objective. Specifically, we show that \(WCSS_w\) reduces to the inverse harmonic sum of individual feature dispersions. This result reveals that the objective is not merely minimised in an abstract sense, but explicitly driven by the most compact features in the data. Since the harmonic sum is dominated by small values, FIR naturally prioritises features with tight within-cluster structure, while attenuating the influence of noisy or weakly informative ones. This formulation makes the behaviour of FIR fully transparent and highlights its role as a dispersion-sensitive rescaling mechanism.

\begin{lemma}
\label{lemma:harmonic}
$WCSS_w$ equals the inverse harmonic sum of feature dispersions.
\end{lemma}
\begin{proof}
Substituting $\alpha^2$ into \(WCSS_w\) leads to 
\begin{align*}
    WCSS_w &= \sum_{v=1}^m \alpha_v^2 D_v= \sum_{v=1}^m  \left( \frac{1}{D_v \sum_{j=1}^m \frac{1}{D_j}} \right)^2 D_v\\
    &=\sum_{v=1}^m \frac{1}{D_v} \left( \frac{1}{\sum_{j=1}^m \frac{1}{D_j}} \right)^2\\
    &=\left( \frac{1}{\sum_{j=1}^m \frac{1}{D_j}} \right)^2 \sum_{v=1}^m \frac{1}{D_v}.
\end{align*}
Clearly, \(\sum_{v=1}^m \frac{1}{D_v}=\sum_{j=1}^m \frac{1}{D_j}\). Hence,
\[
\left( \frac{1}{\sum_{j=1}^m \frac{1}{D_j}} \right)^2 \sum_{v=1}^m \frac{1}{D_v}
=\frac{1}{\sum_{j=1}^m \frac{1}{D_j}}.
\]
\end{proof}

A desirable property of any clustering criterion is robustness to irrelevant or noisy features. In the case of FIR, this corresponds to ensuring that features with arbitrarily high dispersion do not meaningfully affect the objective. The following theorem confirms that FIR satisfies this property. That is, the value of \(WCSS_w\) remains asymptotically unchanged when such features are added.

\begin{theorem}
    \label{thm:unaffected_by_noise}
    \(WCSS_w\) is asymptotically unaffected by the addition of arbitrarily noisy features.    
\end{theorem}
\begin{proof}
    Let us add a new noisy feature to a data set so it contains features \(\{1, \dots, m+1\} \). Then,
    \[
    WCSS_w = \frac{1}{\sum_{j=1}^{m+1} \frac{1}{D_j}} = \frac{1}{\left(\sum_{j=1}^{m} \frac{1}{D_j}\right) + \frac{1}{D_{m+1}}}.
    \]
    But as \( D_{m+1} \to \infty \), we have \( \frac{1}{D_{m+1}} \to 0 \), so:
    \[
    \lim_{D_{m+1} \to \infty} \sum_{v=1}^{m+1} \alpha_v^2 D_v = \frac{1}{\sum_{j=1}^{m} \frac{1}{D_j}},
    \]
    matching the result in Lemma \ref{lemma:harmonic}.
\end{proof}

It is important to note that Theorem \ref{thm:unaffected_by_noise} does not assume independence between features. The result follows solely from the fact that a feature with dispersion $D_{m+1}\to\infty$ contributes $1/D_{m+1}\to 0$ to the harmonic sum, and therefore vanishes from the objective regardless of its dependence structure with other features. This theorem should thus be understood as an asymptotic guarantee for truly uninformative features.

We now turn to the effect of feature scaling. In many real-world applications, features may be measured in different units or undergo rescaling as part of preprocessing. It is therefore important that the method behaves consistently under such transformations. The following proposition shows that FIR satisfies this property: while the weighted objective \(WCSS_w\) is scale dependent, the feature factors \(\alpha_v\) are invariant under uniform scaling of the input features.

\begin{proposition}
    Although the weighted objective \(WCSS_w = \sum_{v=1}^m \alpha_v^2 D_v\) is not scale invariant, each FIR factor \(\alpha_v\) is. In particular, if all dispersions are scaled by a constant factor \(\gamma > 0\), then:
    \[
    D_v^\prime = \sum_{l=1}^k \sum_{x_i \in S_l} (\gamma x_{iv} - \gamma z_{lv})^2 = \gamma^2 D_v.    
    \]
    Thus,
    \[
    \alpha_v' = \frac{1}{\sum_{j=1}^m \frac{\gamma^2 D_v}{\gamma^2 D_j}} = \alpha_v.
    \]
    Hence, FIR behaves identically under uniform feature rescaling.
\end{proposition}

To better understand how FIR responds to feature noise, we examine the sensitivity of the factors \(\alpha_v\) to changes in dispersion. The following proposition shows that \(\alpha_v\) is strictly decreasing in \(D_v\), confirming that FIR down-weights features as their within-cluster dispersion increases. This formalises FIR's role of attenuating the influence of noisy dimensions.

\begin{proposition}
    The FIR factor \(\alpha_v\) is strictly decreasing in \(D_v\). Its sensitivity to changes in dispersion is given by:
    \[
        \frac{\partial \alpha_v}{\partial D_v} = -\frac{1}{\sum_{j=1}^m \frac{D_v^2}{D_j}} \left( 1 - \frac{1}{\sum_{j=1}^m \frac{D_v}{D_j}} \right).
    \]
\end{proposition}
\begin{proof}
We have:
\[
\alpha_v = \frac{1}{\sum_{j=1}^m \frac{D_v}{D_j}} = \frac{1}{D_v} \cdot \left( \frac{1}{\sum_{j=1}^m \frac{1}{D_j}} \right).
\]
Let us differentiate \(\alpha_v\) with respect to \(D_v\):
\begin{align*}
\hspace*{0pt}\frac{\partial \alpha_v}{\partial D_v}
&= -\frac{1}{D_v^2} \cdot \left( \frac{1}{\sum_{j=1}^m \frac{1}{D_j}} \right)\\
&\quad + \frac{1}{D_v} \cdot \left( -\frac{1}{\left(\sum_{j=1}^m \frac{1}{D_j}\right)^2} \cdot \frac{\partial}{\partial D_v} \sum_{j=1}^m \frac{1}{D_j} \right).
\end{align*}
Now observe that:
\[
\frac{\partial}{\partial D_v} \sum_{j=1}^m \frac{1}{D_j} = -\frac{1}{D_v^2}.
\]
Substituting this, we get:
\begin{align*}
\frac{\partial \alpha_v}{\partial D_v} &= -\frac{1}{D_v^2 \sum_{j=1}^m \frac{1}{D_j}} 
+ \frac{1}{D_v} \cdot \left( \frac{1}{\left(\sum_{j=1}^m \frac{1}{D_j}\right)^2} \cdot \frac{1}{D_v^2} \right) \\
&= -\frac{1}{\sum_{j=1}^m \frac{D_v^2}{D_j}} \left( 1 - \frac{1}{\sum_{j=1}^m \frac{D_v}{D_j}} \right).
\end{align*}
Since all terms are positive and the expression is negative, we conclude that \(\alpha_v\) is strictly decreasing in \(D_v\).
\end{proof}

The richness axiom states that for every non-trivial clustering \(S\) of a data set \(X\), there must exist a parameter setting such that \(S\) is the optimal clustering under the corresponding quality function. That is, every possible partition of a data set must be achievable by some parameter configuration \cite{kleinberg2002impossibility}. While this may seem like a natural requirement it can lead to undesirable outcomes, allowing arbitrary or degenerate clusterings. In practice, many effective clustering methods violate richness on purpose to enforce meaningful structure. FIR does not satisfy richness, favouring clusterings that emphasise low-dispersion features.

\begin{theorem}
The clustering quality measure \(WCSS_w\) used by FIR does not satisfy the richness axiom.
\end{theorem}
\begin{proof}
Let us construct a counterexample. Consider the one–dimensional dataset $X=\{0,0,M\}\subset\mathbb{R}$ with $M>0$ and $k=2$. Since $m=1$, the FIR constraint $\sum_{v=1}^m \alpha_v=1$ forces $\alpha_1=1$, so $WCSS_w$ coincides with the ordinary within–cluster sum of squares (WCSS) in this case.

Compare the two $k$-partitions
\[
S_{\text{bad}}=\big\{\{0,M\},\{0\}\big\}
\quad\text{and}\quad
S_{\text{good}}=\big\{\{0,0\},\{M\}\big\}.
\]
For $S_{\text{bad}}$, the centroid of $\{0,M\}$ is $M/2$, hence
\[
WCSS_w(S_{\text{bad}})=\Bigl(0-\tfrac{M}{2}\Bigr)^2+\Bigl(M-\tfrac{M}{2}\Bigr)^2+0
=\tfrac{M^2}{2}.
\]
Note that \(S_{\text{bad}}\) is unstable under $k$-means, at least one point would change its cluster assignment in the next update step. For $S_{\text{good}}$, each cluster is either a singleton or contains identical points, so
\[
WCSS_w(S_{\text{good}})=0.
\]
Therefore $WCSS_w(S_{\text{good}}) < WCSS_w(S_{\text{bad}})$ for every $M>0$. It follows that the partition $S_{\text{bad}}$ is a valid partition but never optimal under FIR. This provides a constructive counterexample, showing that FIR violates the richness axiom.
\end{proof}

%Old (maybe circular) proof
%\begin{proof}
%In the context of FIR, the parameters to evaluate richness are the feature-wise dispersions \(D_1, \ldots, D_m\), which are used to compute feature factors. Lemma \ref{lemma:harmonic} shows that FIR minimises
%\[
%WCSS_w = \frac{1}{\sum_{j=1}^m \frac{1}{D_j}},
%\]
%which is a function solely of the dispersions \(D_1, \ldots, D_m\). However, each dispersion \(D_v\) is defined with respect to a given clustering \(S\); it measures the within-cluster variation of feature \(v\) under \(S\). Thus, the quality measure \(WCSS_w\) is entirely determined by the clustering itself — \(D_v\) cannot be independently specified to favour a given clustering.
%
%Now consider a clustering \(S^*\) whose separation relies primarily on features with high within-cluster dispersion (i.e., large \(D_v\)). These features contribute relatively little to the harmonic sum \( \sum_{j=1}^m \frac{1}{D_j} \), resulting in a smaller denominator and thus a higher value of \(WCSS_w\). Consequently, FIR will penalise such clusterings.
%
%Suppose we attempt to make \(S^*\) optimal by adjusting \(D_1, \ldots, D_m\). To do so, we would need to reduce the values of \(D_v\) for the high-dispersion features, but doing so changes the definition of \(S^*\) itself, since \(D_v\) is a property of the clustering. Therefore, we cannot independently choose \(D_1, \ldots, D_m\) to force \(S^*\) to be optimal — the dependency is circular. Hence, FIR violates the richness axiom.
%\end{proof}

The results in this section provide a clear theoretical foundation for FIR. We have shown that the method is well-posed, interpretable, and robust to irrelevant features. FIR emphasises low-dispersion features through a principled harmonic weighting scheme, and its sensitivity and scale invariance reinforce its practical stability. While FIR violates the richness axiom, this is a deliberate and desirable trade-off that prevents reaching arbitrary or noisy clusterings. These properties explain both the method's internal behaviour and its empirical effectiveness observed in our experiments (see Section \ref{sec:results}).

\section{Setting of the Experiments}
Our primary objective is to fairly evaluate the effectiveness of each of the indices we experiment with (for details, see Sections \ref{sec:related_work} and \ref{sec:new_method}). We achieve this by assessing how well they correlate (or inversely correlate, depending on the index) with the ground truth, despite not being provided with it.

To do so, we first measure cluster recovery using the Adjusted Rand Index (ARI) \cite{hubert1985comparing}, a popular corrected-for-chance version of the Rand Index. We conduct 200 independent runs of $k$-means++, computing the ARI for each clustering outcome against the ground truth. This results in an ARI vector with 200 components. For each of the 200 $k$-means++ runs, we also compute the values of the investigated indices (WCSS, ASW, CH, DB, and their FIR versions - none requiring the ground truth), leading to a separate 200-component vector for each index. Finally, we measure the correlation between the ARI vector and each index vector to evaluate its alignment with the ground truth.

\subsection{Synthetic data sets}
We created a total of 9 basic data configurations, denoted using the notation $n\times m-k$. That is, the configuration $5000\times 20-10$ contains data sets with 5,000 data points, each described over 20 features, and partitioned into 10 clusters. Each data set was generated using \texttt{sklearn.datasets.make\_blobs}, where data points were sampled from a mixture of $k$ Gaussian distributions. More specifically, each cluster $C_l \in C$ follows a multivariate normal distribution: 
\begin{align*}
    x_i \sim \mathcal{N}(z_l, \sigma^2 I),
\end{align*}
where $\sigma$ is the standard deviation controlling the cluster dispersion. For each configuration, we generated eight variations by adding none, $m/2$ (leading to about 33\% noise features in the data set), $m$ (leading to 50\% noise features in the data set), or $4m$ (leading to 80\% noise features in the data set) noise features to the data set, composed of uniformly random values, and setting the cluster dispersion $\sigma$ to either one or two. A value of two leads to more spread clusters, increasing cluster overlap.

In total, we generated 72 unique configurations. Since we created 50 data sets for each configuration, this resulted in a total of 3,600 data sets. We then applied the range (i.e. min-max) normalisation:
\begin{equation}
    \label{eq:range_normalisation}
    x_{iv} = \frac{x_{iv} -\bar{x}_v}{\text{max}\{x_v\}-\text{min}\{x_v\}},
\end{equation}
where $\bar{x}_v$ is the average over all $x_{iv} \in X$, before applying $k$-means++.

\section{Results and discussion}
\label{sec:results}
In this section, we evaluate the impact our data rescaling method has on four internal clustering validation measures: Average Silhouette Width (ASW), Calinski-Harabasz (CH), Davies-Bouldin (DB), and the WCSS in (\ref{eq:kmeans}). More specifically, we assess whether rescaling enhances the correlation between these measures and the ground truth, thereby improving their reliability in unsupervised settings. To this end, we conduct extensive experiments using synthetic data sets with varying feature relevance and cluster structures. By comparing clustering outcomes before and after rescaling, we demonstrate that our method consistently improves the alignment between internal validation indices and external clustering quality measures (ARI), reinforcing its potential to refine clustering evaluation in the absence of labelled data.

It is important to emphasise that our claims are not based on internal indices in isolation. While internal validity indices can be biased, particularly in high-dimensional or imbalanced settings, our evaluation explicitly uses external ground truth (via the ARI) as a reference. The role of FIR is therefore not to improve the clustering quality in itself, but to improve the alignment between internal indices and an external benchmark. By design, our experiments provide access to ground truth labels, allowing us to measure this alignment directly and objectively.

\begin{figure*}[hpt!]
    \centering
    \subfigure[2000x10-5 (PCA)]{\includegraphics[width=0.24\textwidth]{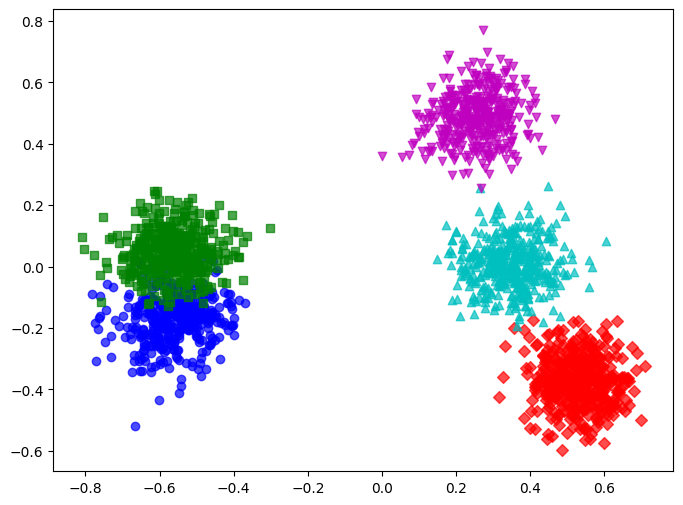}}
    \subfigure[2000x10-5 5NF (PCA)]{\includegraphics[width=0.24\textwidth]{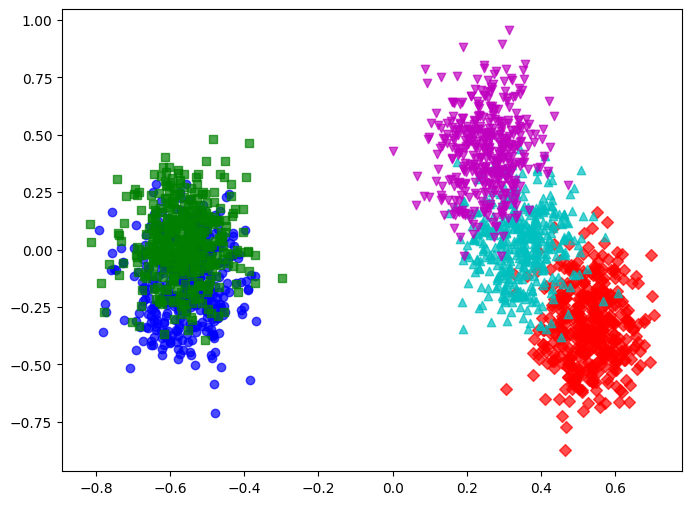}}
    \subfigure[2000x10-5 10NF (PCA)]{\includegraphics[width=0.24\textwidth]{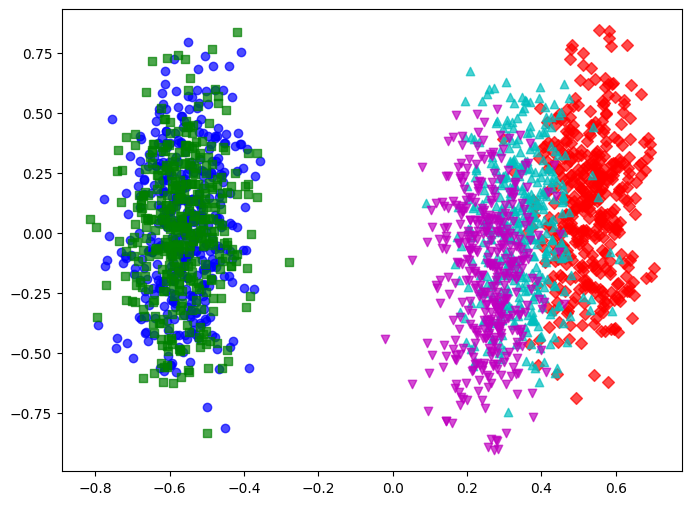}}
    \subfigure[2000x10-5 10NF after FIR (PCA)]{\includegraphics[width=0.24\textwidth]{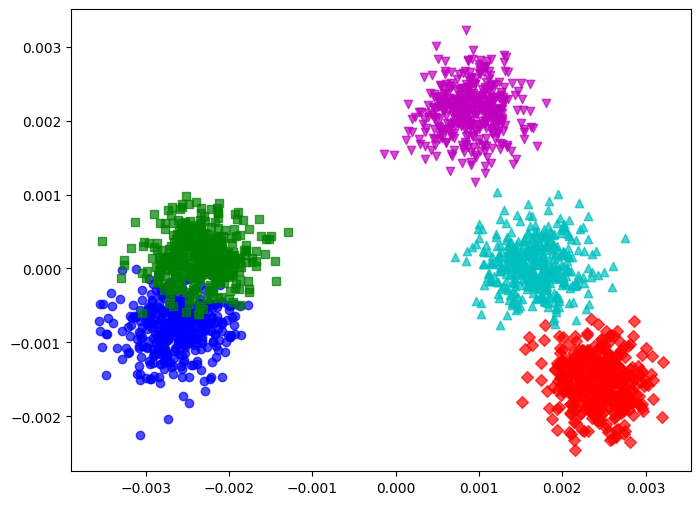}}
    \subfigure[2000x10-5 (T-SNE)]{\includegraphics[width=0.24\textwidth]{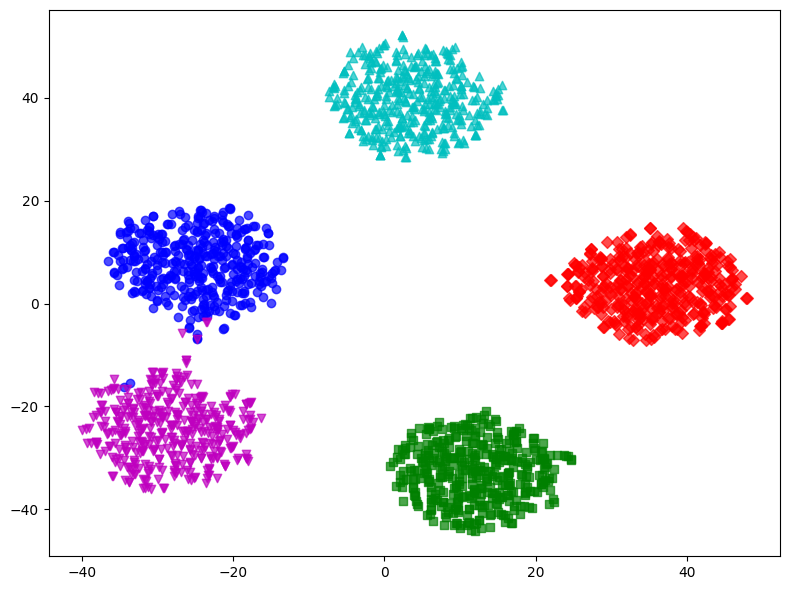}}
    \subfigure[2000x10-5 5NF (T-SNE)]{\includegraphics[width=0.24\textwidth]{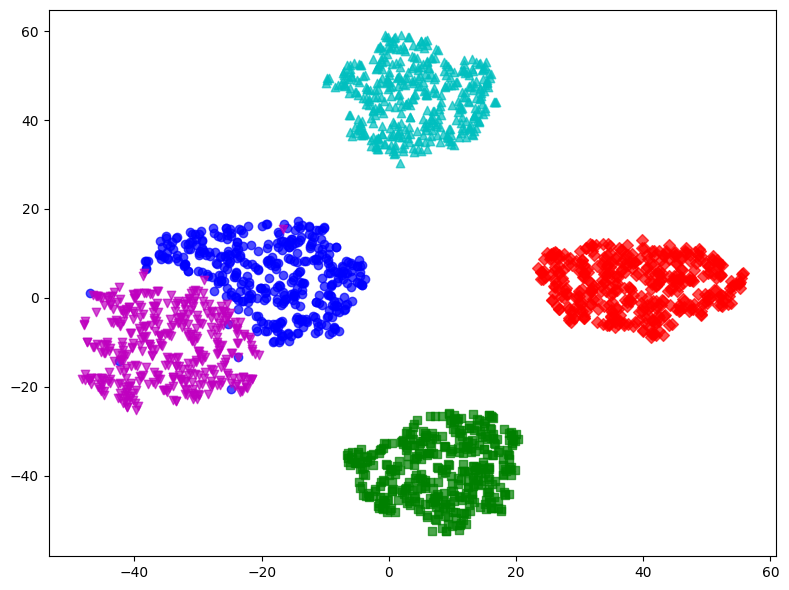}}
    \subfigure[2000x10-5 10NF (T-SNE)]{\includegraphics[width=0.24\textwidth]{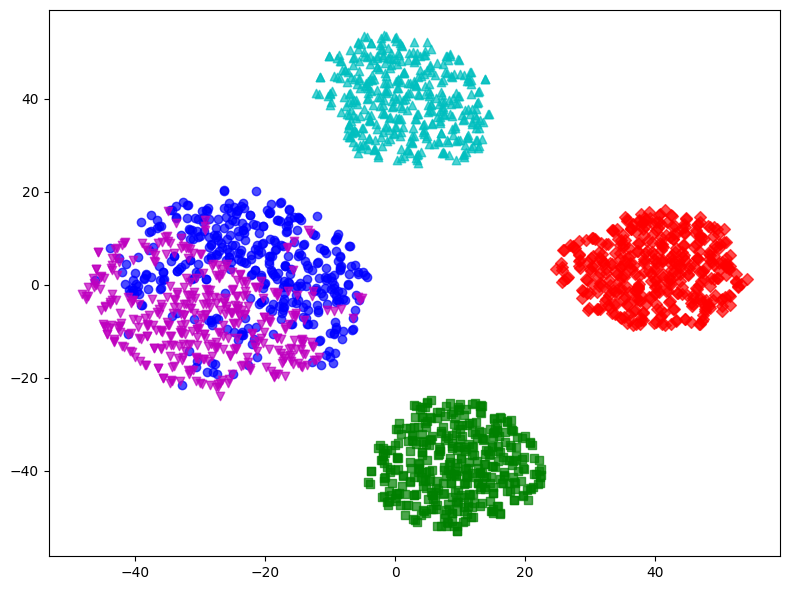}}
    \subfigure[2000x10-5 10NF after FIR (T-SNE)]{\includegraphics[width=0.24\textwidth]{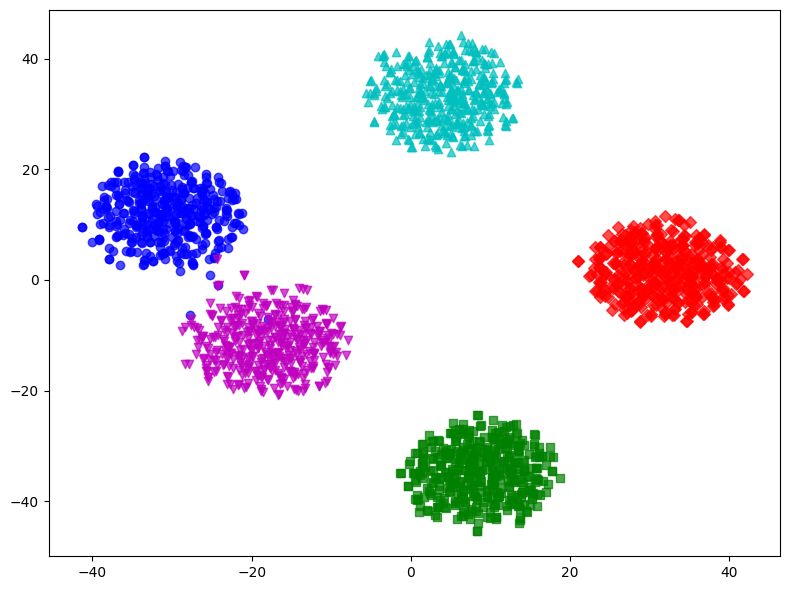}}
    \caption{Projection of the data sets onto their first two principal components after applying PCA, and t-SNE. (a)/(e) Original dataset with 2000 points and 10 features across 5 clusters; (b)/(f) the same data set with five additional noise features; (c)/(g) the dataset with 10 additional noise features; (d)/(h) the dataset with 10 additional noise features after applying our rescaling method.}
    \label{fig:figures}
\end{figure*}

Figure \ref{fig:figures} illustrates the impact of adding noise features on the separability of clusters in a data set, and demonstrates how applying Feature Importance Rescaling (FIR) can improve clustering evaluation in noisy data sets. Subfigure (a) shows an original dataset with 2,000 data points, 10 features, and 5 clusters projected onto the first two principal components using PCA (Principal Component Analysis). In subfigures (b) and (c), we observe that adding 5 and 10 noise features, respectively, creates overlap between clusters, making them less distinguishable in the PCA space. Subfigures (e)-(g) show the same general pattern when using t-SNE \cite{van2008visualizing} instead of PCA. This confirms that introducing irrelevant features complicates the clustering evaluation process by reducing the discriminative power of meaningful dimensions. Finally, subfigures (d) and (h) (for PCA and t-SNE, respectively) present the data set with 10 noise features after applying FIR, showing that clusters become more distinguishable despite the presence of noise. This demonstrates that FIR effectively mitigates the negative impact of noise features by enhancing the separability of clusters, leading to improved clustering performance.

Recall that our goal is not to compare different cluster validity indices to determine which one is superior, but to improve their overall capacity to recover correct clusterings. Extensive studies have already addressed such comparisons (see, for instance, \cite{arbelaitz2013extensive,todeschini2024extended} and references therein). Instead, our objective is to demonstrate that, regardless of which index performs best, our method can further enhance its effectiveness.

Table \ref{tab:Synthetic_1000} presents the average correlation of each index with the ground truth across data sets containing 1,000 data points with varying numbers of features and noise features. The results demonstrate that FIR consistently improves the indices we evaluate, with the most substantial gains observed in data sets containing noise features. This highlights FIR's ability to enhance the robustness of cluster validity measures against irrelevant features. We can also see that FIR improves results even when $\sigma =2$, indicating its effectiveness in scenarios with greater cluster overlap.

\begin{table*}[hpt!]\small
\caption{Experiments on data sets containing 1,000 data points. There are 50 data sets per configuration. For each data set, $k$-means++ was executed 200 times, generating a 200-component ARI vector and a corresponding 200-component vector for each index. The reported correlation measures the alignment between these index vectors and the ARI vector. This number is followed by the standard deviation. Columns labeled ``FIR'' represents results obtained using our proposed method.}
\begin{center}
\tabcolsep=0.08cm
\makebox[\textwidth][c]{%
\begin{tabular}{c lcc|cc|cc|cc}
\toprule
&&WCSS&FIR+WCSS&ASW&FIR+ASW&CH&FIR+CH&DB&FIR+DB\\
%Clusters with a standard deviation of 1, and data points distributed uniformly among clusters 
\midrule 
\multirow{9}{*}{\rotatebox{90}{$\sigma = 1$}}
&1000x6-3&\textbf{-1.00}/0.00&\textbf{-1.00}/0.00&\textbf{1.00}/0.00&\textbf{1.00}/0.00&\textbf{1.00}/0.00&\textbf{1.00}/0.00&\textbf{-1.00}/0.00&\textbf{-1.00}/0.00\\
&1000x6-3 3NF&-0.99/0.02&\textbf{-1.00}/0.01&0.97/0.08&\textbf{0.99}/0.03&\textbf{1.00}/0.02&\textbf{1.00}/0.00&-0.97/0.06&\textbf{-1.00}/0.00\\
&1000x6-3 6NF&-0.98/0.08&\textbf{-1.00}/0.00&0.94/0.21&\textbf{0.98}/0.05&0.98/0.08&\textbf{1.00}/0.00&-0.97/0.11&\textbf{-1.00}/0.01\\
%%%%%%%%%%%%%%%%%
&1000x6-3 24NF&\textbf{-0.98}/0.01&\textbf{-0.98/0.05}&0.76/0.56&\textbf{0.95}/0.27&0.99/0.01&\textbf{1.00}/0.01&-0.95/0.26&\textbf{-0.96}/0.23\\
%%%%%%%%%%%%%%%%%%%
\cmidrule{2-10}
&1000x10-10&\textbf{-0.99}/0.01&\textbf{-0.99}/0.01&\textbf{1.00}/0.00&\textbf{1.00}/0.01&\textbf{1.00}/0.00&\textbf{1.00}/0.01&\textbf{-1.00}/0.00&-0.99/0.01\\
&1000x10-10 5NF&-0.89/0.06&\textbf{-0.96}/0.02&0.84/0.11&\textbf{0.95}/0.04&0.90/0.05&\textbf{0.96}/0.02&-0.76/0.14&\textbf{-0.93}/0.07\\
&1000x10-10 10NF&-0.89/0.06&\textbf{-0.95}/0.03&0.84/0.10&\textbf{0.95}/0.03&0.89/0.06&\textbf{0.94}/0.02&-0.76/0.14&\textbf{-0.94}/0.04\\
%%%%%%%%%%%%%%%%%%%%
&{1000x10-10 40NF}&{\textbf{-0.89}/0.06}&{-0.83/0.17}&{0.82/0.08}&{\textbf{0.95}/0.03}&{0.89/0.06}&{\textbf{0.95}/0.02}&{-0.47/0.16}&{\textbf{-0.90}/0.05}\\
%%%%%%%%%%%%%%%%%
\cmidrule{2-10}
&1000x20-30&\textbf{-0.99}/0.01&\textbf{-0.99}/0.00&\textbf{1.00}/0.00&\textbf{1.00}/0.00&\textbf{0.99}/0.00&\textbf{0.99}/0.01&\textbf{-0.99}/0.00&\textbf{-0.99}/0.01\\
&1000x20-30 10NF&-0.93/0.02&\textbf{-0.96}/0.01&0.92/0.03&\textbf{0.97}/0.01&\textbf{0.93}/0.02&\textbf{0.93}/0.01&-0.88/0.03&\textbf{-0.94}/0.01\\
&1000x20-30 20NF&-0.92/0.02&\textbf{-0.95}/0.01&0.90/0.03&\textbf{0.96}/0.01&0.92/0.02&\textbf{0.93}/0.01&-0.81/0.04&\textbf{-0.93}/0.01\\
%%%%%%%%%%%
&{1000x20-30 80NF}&{-0.97/0.01}&{\textbf{-0.98}/0.00}&{0.90/0.02}&{\textbf{0.97}/0.00}&{\textbf{0.97}/0.01}&{\textbf{0.97}/0.01}&{-0.71/0.07}&{\textbf{-0.95}/0.01}\\
%%%%%%%%%%%%%
\midrule
\multirow{9}{*}{\rotatebox{90}{$\sigma = 2$}}
&1000x6-3&\textbf{-1.00}/0.00&\textbf{-1.00}/0.00&\textbf{0.94}/0.33&\textbf{0.94}/0.33&\textbf{1.00}/0.00&\textbf{1.00}/0.00&\textbf{-1.00}/0.00&\textbf{-1.00}/0.00\\
&1000x6-3 3NF&-0.93/0.15&\textbf{-1.00}/0.01&\textbf{0.85}/0.47&\textbf{0.85}/0.43&0.94/0.15&\textbf{1.00}/0.01&-0.73/0.46&\textbf{-0.96}/0.14\\
&1000x6-3 6NF&-0.89/0.18&\textbf{-0.98}/0.05&\textbf{0.73}/0.55&\textbf{0.73}/0.58&0.89/0.18&\textbf{0.97}/0.06&-0.65/0.54&\textbf{-0.85}/0.29\\
%%%%%%%%%%%%%%%%%%
&{1000x6-3 24NF}&{-0.93/0.11}&{\textbf{-0.97}/0.05}&{0.59/0.64}&{\textbf{0.60}/0.70}&{0.93/0.11}&{\textbf{0.97}/0.07}&{-0.41/0.75}&{\textbf{-0.58}/0.63}\\
%%%%%%%%%%%%%%%%%%
\cmidrule{2-10}
&1000x10-10&-0.97/0.04&\textbf{-0.98}/0.03&\textbf{0.96}/0.07&\textbf{0.96}/0.08&\textbf{0.98}/0.02&\textbf{0.98}/0.02&\textbf{-0.99}/0.01&\textbf{-0.99}/0.01\\
&1000x10-10 5NF&-0.81/0.10&\textbf{-0.96}/0.03&0.85/0.08&\textbf{0.95}/0.03&0.82/0.10&\textbf{0.95}/0.03&-0.49/0.17&\textbf{-0.86}/0.08\\
&1000x10-10 10NF&-0.88/0.07&\textbf{-0.97}/0.02&0.88/0.07&\textbf{0.96}/0.02&0.88/0.06&\textbf{0.96}/0.02&-0.38/0.24&\textbf{-0.86}/0.07\\
%%%%%%%%%%%%%%%%%%%%
&{1000x10-10 40NF}&{-0.88/0.07}&{\textbf{-0.96}/0.03}&{0.83/0.10}&{\textbf{0.93}/0.05}&{0.88/0.07}&{\textbf{0.95}/0.03}&{-0.19/0.18}&{\textbf{-0.74}/0.15}\\
%%%%%%%%%%%%%%%%%%%
\cmidrule{2-10}
&1000x20-30&-0.96/0.03&\textbf{-0.97}/0.02&\textbf{0.97}/0.03&\textbf{0.97}/0.02&\textbf{0.96}/0.02&\textbf{0.96}/0.02&\textbf{-0.96}/0.01&\textbf{-0.96}/0.01\\
&1000x20-30 10NF&-0.92/0.02&\textbf{-0.96}/0.01&0.91/0.02&\textbf{0.97}/0.01&0.92/0.02&\textbf{0.95}/0.01&-0.78/0.05&\textbf{-0.93}/0.02\\
&1000x20-30 20NF&-0.94/0.01&\textbf{-0.98}/0.01&0.91/0.02&\textbf{0.98}/0.01&0.94/0.01&\textbf{0.97}/0.01&-0.62/0.08&\textbf{-0.93}/0.02\\
%%%%%%%%%%%%%%%%%%%%
&{1000x20-30 80NF}&{-0.68/0.10}&{\textbf{-0.93}/0.02}&{0.35/0.09}&{\textbf{0.50}/0.12}&{0.68/0.10}&{\textbf{0.91}/0.03}&{-0.37/0.09}&{\textbf{-0.62}/0.07}\\
%%%%%%%%%%%%%%%%%%%%
\bottomrule
\end{tabular}}
\end{center}
\label{tab:Synthetic_1000}
\end{table*}

Table \ref{tab:Synthetic_2000} presents the results of similar experiments on data sets with 2,000 data points. The overall pattern closely aligns with that observed in Table \ref{tab:Synthetic_1000}, with FIR consistently enhancing correlation across all indices. As expected, the improvement is most pronounced in noisy scenarios, and remains strong even with a higher degree of overlap between clusters ($\sigma=2$).

\begin{table*}[hpt!]\small
%\caption{The correlation between each index and the ground truth. Experiments with 2000 data points were conducted. There are 50 data sets for each configuration. $K$-means++ was carried out 200 times for each data set. The columns with FIR corresponds to the results obtained using our method.}
\caption{Experiments on data sets containing 2,000 data points. There are 50 data sets per configuration. For each data set, $k$-means++ was executed 200 times, generating a 200-component ARI vector and a corresponding 200-component vector for each index. The reported correlation measures the alignment between these index vectors and the ARI vector. This number is followed by the standard deviation. Columns labeled ``FIR'' represent results obtained using our proposed method.}
\begin{center}
\tabcolsep=0.08cm
\makebox[\textwidth][c]{%
\begin{tabular}{c lcc|cc|cc|cc}
\toprule
&&WCSS&FIR+WCSS&ASW&FIR+ASW&CH&FIR+CH&DB&FIR+DB\\
%Clusters with a standard deviation of 1, and data points distributed uniformly among clusters 
\midrule
\multirow{9}{*}{\rotatebox{90}{$\sigma = 1$}}
&2000x10-5&\textbf{-1.00}/0.01&\textbf{-1.00}/0.01&\textbf{1.00}/0.00&\textbf{1.00}/0.00&\textbf{1.00}/0.00&\textbf{1.00}/0.01&\textbf{-1.00}/0.00&\textbf{-1.00}/0.01\\
&2000x10-5 5NF&-0.96/0.07&\textbf{-0.98}/0.02&0.94/0.14&\textbf{0.98}/0.03&0.97/0.06&\textbf{0.99}/0.01&-0.97/0.08&\textbf{-0.99}/0.02\\
&2000x10-5 10NF&-0.95/0.06&\textbf{-0.98}/0.02&0.90/0.21&\textbf{0.98}/0.03&0.96/0.06&\textbf{0.99}/0.01&-0.97/0.07&\textbf{-0.99}/0.02\\
%%%%%%%%%%%%%%%%%%
&{2000x10-5 40NF}&{\textbf{-0.93}/0.08}&{\textbf{-0.93}/0.20}&{0.81/0.28}&{\textbf{0.96}/0.06}&{0.94/0.08}&{\textbf{0.99}/0.02}&{-0.94/0.09}&{\textbf{-0.99}/0.01}\\
%%%%%%%%%%%%%%%%%
\cmidrule{2-10}
&2000x20-20&\textbf{-0.99}/0.01&\textbf{-0.99}/0.01&\textbf{1.00}/0.00&\textbf{1.00}/0.00&\textbf{0.99}/0.00&\textbf{0.99}/0.01&\textbf{-1.00}/0.00&\textbf{-1.00}/0.00\\
&2000x20-20 10NF&-0.93/0.03&\textbf{-0.96}/0.02&0.92/0.04&\textbf{0.97}/0.01&\textbf{0.93}/0.03&\textbf{0.93}/0.01&-0.94/0.02&\textbf{-0.97}/0.01\\
&2000x20-20 20NF&-0.93/0.03&\textbf{-0.95}/0.02&0.91/0.04&\textbf{0.96}/0.01&\textbf{0.92}/0.03&\textbf{0.92}/0.01&-0.93/0.02&\textbf{-0.97}/0.01\\
%%%%%%%%%%%%%%%%%%
&{2000x20-20 80NF}&{\textbf{-0.92}/0.02}&{-0.32/0.20}&{0.89/0.04}&{\textbf{0.95}/0.01}&{0.92/0.02}&{\textbf{0.94}/0.01}&{-0.76/0.07}&{\textbf{-0.94}/0.02}\\
%%%%%%%%%%%%%%%%%%
\cmidrule{2-10}
&2000x30-40&\textbf{-0.99}/0.01&\textbf{-0.99}/0.00&\textbf{1.00}/0.00&\textbf{1.00}/0.00&\textbf{0.99}/0.00&\textbf{0.99}/0.00&\textbf{-0.99}/0.00&\textbf{-0.99}/0.00\\
&2000x30-40 15NF&\textbf{-0.96}/0.01&\textbf{-0.96}/0.01&0.95/0.02&\textbf{0.97}/0.01&\textbf{0.94}/0.01&\textbf{0.94}/0.01&-0.94/0.01&\textbf{-0.97}/0.01\\
&2000x30-40 30NF&\textbf{-0.95}/0.01&\textbf{-0.95}/0.01&0.94/0.02&\textbf{0.96}/0.01&\textbf{0.93}/0.01&\textbf{0.93}/0.01&-0.92/0.01&\textbf{-0.96}/0.01\\
%%%%%%%%%%%%%%%%%%%
&{2000x30-40 120NF}&{\textbf{-0.95}/0.01}&{\textbf{-0.95}/0.02}&{0.91/0.02}&{\textbf{0.96}/0.01}&{\textbf{0.95}/0.01}&{\textbf{0.95}/0.01}&{-0.72/0.04}&{\textbf{-0.92}/0.01}\\
%%%%%%%%%%%%%%%%%%%
\midrule
\multirow{9}{*}{\rotatebox{90}{$\sigma = 2$}}
&2000x10-5&\textbf{-0.99}/0.02&\textbf{-0.99}/0.02&\textbf{0.97}/0.16&0.96/0.19&\textbf{1.00}/0.01&0.99/0.01&\textbf{-1.00}/0.00&-0.99/0.01\\
&2000x10-5 5NF&-0.92/0.08&\textbf{-0.98}/0.03&0.90/0.14&\textbf{0.94}/0.11&0.93/0.08&\textbf{0.98}/0.02&-0.86/0.22&\textbf{-0.97}/0.05\\
&2000x10-5 10NF&-0.93/0.09&\textbf{-0.98}/0.04&0.87/0.21&\textbf{0.91}/0.23&0.93/0.08&\textbf{0.98}/0.03&-0.84/0.26&\textbf{-0.95}/0.13\\
%%%%%%%%%%%%%%%%%%%
&{2000x10-5 40NF}&{-0.93/0.07}&{\textbf{-0.98}/0.02}&{0.90/0.15}&{\textbf{0.93}/0.15}&{0.93/0.07}&{\textbf{0.98}/0.02}&{-0.76/0.31}&{\textbf{-0.92}/0.19}\\
%%%%%%%%%%%%%%%%%%%%%
\cmidrule{2-10}
&2000x20-20&-0.97/0.03&\textbf{-0.98}/0.02&\textbf{0.98}/0.02&\textbf{0.98}/0.02&\textbf{0.97}/0.02&\textbf{0.97}/0.02&\textbf{-0.99}/0.01&\textbf{-0.99}/0.01\\
&2000x20-20 10NF&-0.92/0.03&\textbf{-0.95}/0.01&0.91/0.04&\textbf{0.96}/0.01&0.92/0.03&\textbf{0.94}/0.01&-0.89/0.04&\textbf{-0.97}/0.01\\
&2000x20-20 20NF&-0.91/0.03&\textbf{-0.95}/0.01&0.90/0.04&\textbf{0.96}/0.01&0.92/0.03&\textbf{0.95}/0.02&-0.82/0.07&\textbf{-0.95}/0.02\\
%%%%%%%%%%%%%%%%%%%
&{2000x20-20 80NF}&{-0.98/0.01}&{\textbf{-0.99}/0.00}&{0.95/0.03}&{\textbf{0.97}/0.01}&{\textbf{0.98}/0.01}&{\textbf{0.98}/0.01}&{-0.52/0.14}&{\textbf{-0.95}/0.02}\\
%%%%%%%%%%%%%%%%%%%
\cmidrule{2-10}
&2000x30-40&-0.97/0.01&\textbf{-0.98}/0.01&\textbf{0.98}/0.01&\textbf{0.98}/0.01&\textbf{0.97}/0.01&\textbf{0.97}/0.01&-0.96/0.01&\textbf{-0.97}/0.01\\
&2000x30-40 15NF&-0.94/0.01&\textbf{-0.96}/0.01&0.94/0.02&\textbf{0.97}/0.01&0.94/0.01&\textbf{0.95}/0.01&-0.91/0.02&\textbf{-0.96}/0.01\\
&2000x30-40 30NF&-0.94/0.01&\textbf{-0.95}/0.01&0.93/0.02&\textbf{0.96}/0.01&0.94/0.01&\textbf{0.95}/0.01&-0.84/0.03&\textbf{-0.94}/0.01\\
%%%%%%%%%%%%%%%%%%%%%%
&{2000x30-40 120NF}&{-0.96/0.01}&{\textbf{-0.99}/0.00}&{0.79/0.08}&{\textbf{0.92}/0.04}&{0.96/0.01}&{\textbf{0.99}/0.00}&{-0.77/0.06}&{\textbf{-0.92}/0.03}\\
%%%%%%%%%%%%%%%%%%%%%
\bottomrule
\end{tabular}}
\end{center}
\label{tab:Synthetic_2000}
\end{table*}

Table \ref{tab:Synthetic_5000} presents the results of similar experiments on data sets with 5,000 data points. Once again, the overall pattern aligns with those observed in Tables \ref{tab:Synthetic_1000} and \ref{tab:Synthetic_2000}, with FIR consistently improving the correlation across all indices. Notably, the impact of FIR on the DB index is more pronounced in this setting. Additionally, it is interesting to observe that experiments with a larger number of data points generally exhibit lower standard deviations, suggesting increased stability in the results.

\begin{table*}[hpt!]\small
%\caption{The correlation between each index and the ground truth. Experiments with 5000 data points were conducted. There are 50 data sets for each configuration. $K$-means++ was carried out 200 times for each data set. The columns with FIR corresponds to the results obtained using our method.}
\caption{Experiments on data sets containing 5,000 data points. There are 50 data sets per configuration. For each data set, $k$-means++ was executed 200 times, generating a 200-component ARI vector and a corresponding 200-component vector for each index. The reported correlation measures the alignment between these index vectors and the ARI vector. This number is followed by the standard deviation. Columns labeled ``FIR'' represent results obtained using our proposed method.}
\begin{center}
\tabcolsep=0.08cm
\makebox[\textwidth][c]{%
\begin{tabular}{c lcc|cc|cc|cc}
\toprule
&&WCSS&FIR+WCSS&ASW&FIR+ASW&CH&FIR+CH&DB&FIR+DB\\
%Clusters with a standard deviation of 1, and data points distributed uniformly among clusters 
\midrule
\multirow{9}{*}{\rotatebox{90}{$\sigma = 1$}}
&5000x20-10&\textbf{-1.00}/0.00&\textbf{-1.00}/0.00&\textbf{1.00}/0.00&\textbf{1.00}/0.00&\textbf{1.00}/0.00&\textbf{1.00}/0.00&\textbf{-1.00}/0.00&\textbf{-1.00}/0.00\\
&5000x20-10 10NF&-0.96/0.03&\textbf{-0.98}/0.01&0.95/0.04&\textbf{0.99}/0.01&0.97/0.03&\textbf{0.98}/0.01&\textbf{-0.99}/0.01&\textbf{-0.99}/0.00\\
&5000x20-10 20NF&-0.95/0.03&\textbf{-0.97}/0.01&0.93/0.05&\textbf{0.98}/0.01&0.95/0.03&\textbf{0.96}/0.01&-0.98/0.01&\textbf{-0.99}/0.00\\
%%%%%%%%%%%%%%%%%%
&{5000x20-10 80NF}&{\textbf{-0.94}/0.03}&{-0.41/0.22}&{0.92/0.06}&{\textbf{0.96}/0.02}&{0.94/0.03}&{\textbf{0.95}/0.01}&{-0.96/0.02}&{\textbf{-0.98}/0.01}\\
%%%%%%%%%%%%%%%%%%%
\cmidrule{2-10}
&5000x30-30&\textbf{-1.00}/0.00&\textbf{-1.00}/0.00&\textbf{1.00}/0.00&\textbf{1.00}/0.00&\textbf{1.00}/0.00&\textbf{1.00}/0.00&\textbf{-1.00}/0.00&\textbf{-1.00}/0.00\\
&5000x30-30 15NF&-0.96/0.01&\textbf{-0.97}/0.01&0.96/0.02&\textbf{0.97}/0.01&\textbf{0.94}/0.01&\textbf{0.94}/0.01&\textbf{-0.97}/0.01&\textbf{-0.97}/0.01\\
&5000x30-30 30NF&\textbf{-0.95}/0.01&\textbf{-0.95}/0.01&0.94/0.02&\textbf{0.97}/0.01&\textbf{0.94}/0.01&\textbf{0.94}/0.01&-0.96/0.01&\textbf{-0.97}/0.01\\
%%%%%%%%%%%%%%%%%%%%%
&{5000x30-30 120NF}&{\textbf{-0.93}/0.02}&{-0.46/0.12}&{0.91/0.03}&{\textbf{0.95}/0.01}&{\textbf{0.93}/0.02}&{\textbf{0.93}/0.01}&{-0.89/0.03}&{\textbf{-0.95}/0.01}\\
%%%%%%%%%%%%%%%%%%%%%
\cmidrule{2-10}
&5000x40-50&-0.99/0.00&\textbf{-1.00}/0.00&\textbf{1.00}/0.00&\textbf{1.00}/0.00&\textbf{0.99}/0.00&\textbf{0.99}/0.00&\textbf{-0.99}/0.00&\textbf{-0.99}/0.00\\
&5000x40-50 20NF&\textbf{-0.96}/0.01&\textbf{-0.96}/0.01&0.95/0.01&\textbf{0.97}/0.01&\textbf{0.95}/0.01&\textbf{0.95}/0.01&-0.96/0.01&\textbf{-0.97}/0.01\\
&5000x40-50 40NF&\textbf{-0.95}/0.01&\textbf{-0.95}/0.01&0.95/0.01&\textbf{0.96}/0.01&\textbf{0.94}/0.01&\textbf{0.94}/0.01&-0.95/0.01&\textbf{-0.97}/0.01\\
%%%%%%%%%%%%%%%%%
&{5000x40-50 160NF}&{\textbf{-0.93}/0.01}&{-0.89/0.05}&{0.92/0.02}&{\textbf{0.94}/0.01}&{\textbf{0.93}/0.01}&{\textbf{0.93}/0.01}&{-0.86/0.02}&{\textbf{-0.93}/0.01}\\
%%%%%%%%%%%%%%%%
\midrule
\multirow{9}{*}{\rotatebox{90}{$\sigma = 2$}}
&5000x20-10&\textbf{-0.99}/0.01&\textbf{-0.99}/0.01&\textbf{0.99}/0.01&\textbf{0.99}/0.01&\textbf{0.99}/0.01&\textbf{0.99}/0.01&\textbf{-1.00}/0.00&\textbf{-1.00}/0.00\\
&5000x20-10 10NF&-0.94/0.03&\textbf{-0.97}/0.02&0.93/0.06&\textbf{0.98}/0.02&0.95/0.03&\textbf{0.97}/0.02&-0.97/0.02&\textbf{-0.99}/0.01\\
&5000x20-10 20NF&-0.94/0.03&\textbf{-0.97}/0.02&0.93/0.05&\textbf{0.97}/0.02&0.95/0.03&\textbf{0.96}/0.01&-0.97/0.02&\textbf{-0.99}/0.00\\
%%%%%%%%%%%%%%%%%%%%%
&{5000x20-10 80NF}&{-0.93/0.04}&{\textbf{-0.96}/0.02}&{0.89/0.08}&{\textbf{0.96}/0.03}&{0.93/0.04}&{\textbf{0.97}/0.02}&{-0.92/0.06}&{\textbf{-0.99}/0.01}\\
%%%%%%%%%%%%%%%%%%%%%
\cmidrule{2-10}
&5000x30-30&\textbf{-0.98}/0.01&\textbf{-0.98}/0.01&\textbf{0.99}/0.01&\textbf{0.99}/0.01&\textbf{0.98}/0.01&\textbf{0.98}/0.01&\textbf{-0.99}/0.01&\textbf{-0.99}/0.00\\
&5000x30-30 15NF&-0.95/0.02&\textbf{-0.96}/0.01&0.95/0.02&\textbf{0.97}/0.01&0.95/0.02&\textbf{0.96}/0.01&-0.95/0.01&\textbf{-0.98}/0.01\\
&5000x30-30 30NF&-0.94/0.02&\textbf{-0.96}/0.01&0.94/0.02&\textbf{0.96}/0.01&0.94/0.02&\textbf{0.95}/0.01&-0.94/0.01&\textbf{-0.97}/0.01\\
%%%%%%%%%%%%%%%%%
&{5000x30-30 120NF}&{-0.95/0.01}&{\textbf{-0.97}/0.01}&{0.91/0.02}&{\textbf{0.97}/0.01}&{0.95/0.01}&{\textbf{0.96}/0.01}&{-0.59/0.08}&{\textbf{-0.92}/0.02}\\
%%%%%%%%%%%%%%%%%
\cmidrule{2-10}
&5000x40-50&\textbf{-0.98}/0.01&\textbf{-0.98}/0.01&\textbf{0.98}/0.01&\textbf{0.98}/0.01&0.97/0.01&\textbf{0.98}/0.01&\textbf{-0.97}/0.01&\textbf{-0.97}/0.01\\
&5000x40-50 20NF&-0.95/0.01&\textbf{-0.96}/0.01&0.95/0.01&\textbf{0.97}/0.01&\textbf{0.95}/0.01&\textbf{0.95}/0.01&-0.95/0.01&\textbf{-0.97}/0.01\\
&5000x40-50 40NF&\textbf{-0.95}/0.01&\textbf{-0.95}/0.01&0.94/0.02&\textbf{0.96}/0.01&\textbf{0.95}/0.01&\textbf{0.95}/0.01&-0.93/0.01&\textbf{-0.96}/0.01\\
%%%%%%%%%%%%%%%%%%%%
&{5000x40-50 160NF}&{\textbf{-0.98}/0.01}&{\textbf{-0.98}/0.01}&{0.91/0.03}&{\textbf{0.97}/0.01}&{\textbf{0.98}/0.01}&{\textbf{0.98}/0.01}&{-0.72/0.11}&{\textbf{-0.94}/0.03}\\
%&Malware&0.83/0.07&0.83/0.12&nan/nan&nan/nan&-0.82/0.08&-0.82/0.13&0.73/0.13&0.80/0.14\\
%&Malware&0.87/0.06&0.90/0.08&nan/nan&nan/nan&-0.87/0.07&-0.90/0.10&0.80/0.10&0.87/0.10\\
%&Malware&0.86/0.08&0.88/0.11&nan/nan&nan/nan&-0.85/0.08&-0.88/0.12&0.77/0.12&0.85/0.12\\
%&nomao.ma&-0.99/0.00&-0.97/0.03&nan/nan&nan/nan&0.99/0.00&0.90/0.03&-0.98/0.01&-0.73/0.21\\
%&nomao.ma&-0.99/0.00&-0.98/0.02&nan/nan&nan/nan&0.99/0.00&0.91/0.03&-0.99/0.01&-0.77/0.18\\
%&gas&0.82/0.10&0.86/0.07&nan/nan&nan/nan&-0.83/0.08&-0.87/0.07&-0.12/0.19&0.00/0.20\\
%&har.ma&0.61/0.28&-0.36/0.00&nan/nan&nan/nan&-0.62/0.26&0.28/0.06&0.90/0.05&0.39/0.14\\
%&har.ma&0.66/0.22&nan/nan&-0.10/0.13&0.33/0.17&-0.67/0.20&0.26/0.08&0.91/0.04&0.42/0.14\\
%&har.ma&0.63/0.25&nan/nan&-0.10/0.14&0.32/0.18&-0.64/0.24&0.24/0.09&0.89/0.07&0.42/0.14\\
%&har.ma&0.61/0.28&-0.36/0.00&-0.10/0.13&0.33/0.14&-0.62/0.26&0.28/0.06&0.90/0.05&0.39/0.14\\
%%%%%%%%%%%%%%%%%%%
\bottomrule
\end{tabular}}
\end{center}
\label{tab:Synthetic_5000}
\end{table*}

For each reported mean and standard deviation, we also computed the 95\% confidence intervals. The resulting half-widths are at most $\pm 0.06$, and in the vast majority of cases below $\pm 0.01$. This confirms that the averages reported in Tables~\ref{tab:Synthetic_1000}--\ref{tab:Synthetic_5000} are statistically robust without the need for separate significance testing. Moreover, we note that in the majority of cases the application of FIR reduces the reported standard deviations, leading to tighter confidence intervals and hence more reliable index behaviour.

FIR derives its rescaling factors from the within-cluster dispersions \(D_v\), which are computed with respect to a given clustering solution. A potential concern is therefore whether inaccuracies in the clustering could lead to unstable estimates of feature relevance. In practice, the influence of such effects is limited for two reasons. First, the factors \(\alpha_v\) depend on the relative magnitudes of the dispersions rather than their absolute values, meaning that moderate variations in cluster assignments typically produce only small changes in the resulting rescaling. Second, our experimental protocol already evaluates FIR under substantial clustering variability: each data set is clustered 200 times using \(k\)-means++, including scenarios with strong cluster overlap (\(\sigma=2\)) and high proportions of noise features. The consistently improved correlations and the generally reduced standard deviations observed when FIR is applied indicate that the method remains stable and beneficial even when the underlying clustering solutions are relatively weak.

Table~\ref{tab:time} reports the average running time of $k$-means++ with and without FIR across different data set configurations. As expected from the theoretical analysis in Section~\ref{sec:new_method}, incorporating FIR introduces only a marginal increase in computational cost. Across all configurations, the additional overhead remains negligible compared to the overall clustering time, and does not alter the asymptotic behaviour of the algorithm (see also Theorem~\ref{thm:free}). Moreover, the standard deviations of the running times remain essentially unchanged when FIR is applied. This is expected, since FIR consists of deterministic operations whose cost depends only on the data dimensionality and number of points, and does not introduce additional sources of randomness beyond those already present in the $k$-means++ initialisation. These results confirm that FIR acts as a computationally lightweight enhancement in practice, in line with the theoretical guarantees in Section~\ref{sec:new_method}.

\begin{table*}[hpt!]\small
\caption{Average running time in seconds for different dataset configurations. There are 50 data sets per configuration. For each data set, $k$-means++ was executed 200 times. The reported time is followed by the standard deviation. We experimented with $k$-means++ and $k$-means++ followed by FIR.}
\begin{center}
\tabcolsep=0.15cm
\makebox[\textwidth][c]{%
\begin{tabular}{lcc}
\toprule
&{$k$-means++}&{$k$-means++ with FIR}\\
\midrule
{1000x6-3}&{0.0012/0.0009}&{0.0016/0.0009}\\
{1000x10-10}&{0.0085/0.0034}&{0.0091/0.0035}\\
{1000x20-30}&{0.0469/0.0041}&{0.0479/0.0041}\\
{2000x10-5}&{0.0055/0.0045}&{0.0066/0.0045}\\
{2000x20-20}&{0.0526/0.0124}&{0.0544/0.0124}\\
{2000x30-40}&{0.1715/0.0154}&{0.1741/0.0154}\\
{5000x20-10}&{0.0640/0.0461}&{0.0685/0.0461}\\
{5000x30-30}&{0.3510/0.0746}&{0.3574/0.0746}\\
{5000x40-50}&{0.8727/0.0916}&{0.8810/0.0916}\\
\bottomrule
\end{tabular}}
\end{center}
\label{tab:time}
\end{table*}

\subsection{Evaluating clustering-dependent feature rescaling}
This section investigates whether the improvements obtained with FIR arise from its use of clustering structure, or whether similar improvements could be achieved by a simpler global variance-based rescaling of the data. To this end, we compare FIR with inverse-variance normalisation (InvVar), a common unsupervised baseline that rescales features using only global statistics and does not rely on clustering information.

Let $\sigma_v^2$ denote the empirical variance of feature $v$ computed over the entire data set. Inverse-variance normalisation rescales each feature proportionally to the inverse of this variance. That is,
\[
\alpha_v=\frac{\frac{1}{\sigma_v^2}}{\sum_{j=1}^{m}\frac{1}{\sigma_j^2}},
\]
and the data are rescaled according to $x'_{iv}=\alpha_v x_{iv}$. Recall that FIR derives its feature factors from the within-cluster dispersions $D_v$ (see Eq.~\ref{eq:dispersion}), which depend explicitly on the cluster assignments. As a result, FIR incorporates information about how each feature behaves within clusters, rather than relying solely on global variability. InvVar, unlike FIR, depends only on the global variance of each feature and is therefore independent of the clustering structure.

Table~\ref{tab:invvar_comparison} compares the correlation with the ground truth obtained using InvVar and FIR across representative noisy synthetic data sets. The experimental protocol is identical to that used in the previous experiments: for each data set configuration we generated 50 data sets, executed $k$-means++ 200 times per data set, and measured the correlation between each internal index and the ARI with respect to the ground truth. The results show that FIR consistently achieves stronger alignment between the internal indices and the ground truth than InvVar. This indicates that the improvements produced by FIR are not merely a consequence of variance-based rescaling, but arise from the clustering-dependent information captured by $D_v$.

\begin{table*}[hpt!]\small
\caption{Comparison between inverse-variance normalisation (InvVar) and FIR on representative noisy synthetic data sets. There are 50 data sets per configuration. For each data set, $k$-means++ was executed 200 times, generating a 200-component ARI vector and a corresponding 200-component vector for each index. The reported correlation measures the alignment between these index vectors and the ARI vector. This number is followed by the standard deviation.}
\begin{center}
\tabcolsep=0.09cm
\makebox[\textwidth][c]{%
\begin{tabular}{lcccccccc}
\toprule
& \multicolumn{2}{c}{WCSS}
& \multicolumn{2}{c}{ASW} 
& \multicolumn{2}{c}{CH} 
& \multicolumn{2}{c}{DB}\\ 
\cmidrule(r){2-3}
\cmidrule(r){4-5}
\cmidrule(r){6-7}
\cmidrule(r){8-9}
& InvVar & FIR 
& InvVar & FIR 
& InvVar & FIR 
& InvVar & FIR \\
\midrule
1000x6-3 3NF&\textbf{-1.00}/0.01&\textbf{-1.00}/0.0&\textbf{0.99}/0.04&\textbf{0.99}/0.0&\textbf{1.00}/0.01&\textbf{1.00}/0.00&-0.99/0.02&\textbf{-1.00}/0.00\\
1000x6-3 6NF&-0.99/0.01&\textbf{-1.00}/0.00&0.96/0.11&\textbf{0.98}/0.05&\textbf{1.00}/0.01&\textbf{1.00}/0.00&-0.99/0.01&\textbf{-1.00
}/0.01\\
%1000x6-3 24NF&\textbf{-0.98}/0.02&\textbf{-0.98}/0.05&0.92/0.28&\textbf{0.95}/0.27&0.99/0.02&\textbf{1.00}/0.01&-0.99/0.01&-0.96/0.23\\
1000x10-10 5NF&-0.93/0.04&\textbf{-0.96}/0.02&0.91/0.07&\textbf{0.95}/0.04&0.93/0.04&\textbf{0.96}/0.02&-0.91/0.07&\textbf{-0.93}/0.07\\
1000x10-10 10NF&-0.92/0.05&\textbf{-0.95}/0.03&0.88/0.07&\textbf{0.95}/0.03&0.92/0.05&\textbf{0.94}/0.02&-0.90/0.05&\textbf{-0.94}/0.04\\
%1000x10-10 40NF&\textbf{-0.92}/0.04&-0.83/0.17&0.86/0.07&\textbf{0.94}/0.03&0.93/0.05&\textbf{0.95}/0.02&-0.76/0.10&\textbf{-0.90}/0.05\\
1000x20-30 10NF&-0.94/0.02&\textbf{-0.96}/0.01&0.94/0.02&\textbf{0.97}/0.01&\textbf{0.93}/0.02&\textbf{0.93}/0.01&-0.93/0.02&\textbf{-0.94}/0.01\\
1000x20-30 20NF&-0.94/0.02&\textbf{-0.95}/0.01&0.93/0.03&\textbf{0.96}/0.01&\textbf{0.93}/0.02&\textbf{0.93}/0.01&-0.91/0.02&\textbf{-0.93}/0.01\\
%1000x20-30 80NF&-0.94/0.01&\textbf{-0.95}/0.00&0.93/0.01&\textbf{0.96}/0.00&0.92/0.01&\textbf{0.93}/0.01&-0.87/0.04&\textbf{-0.93}/0.01\\
\midrule
2000x10-5 5NF&-0.97/0.04&\textbf{-0.98}/0.02&0.95/0.10&\textbf{0.98}/0.03&0.97/0.03&\textbf{0.99}/0.01&-0.98/0.06&\textbf{-0.99}/0.02\\
2000x10-5 10NF&-0.96/0.04&\textbf{-0.98}/0.02&0.93/0.12&\textbf{0.98}/0.03&0.96/0.04&\textbf{0.99}/0.01&\textbf{-0.99}/0.01&\textbf{-0.99}/0.02\\
%2000x10-5 40NF&\textbf{-0.95}/0.05&-0.93/0.20&0.91/0.14&\textbf{0.96}/0.06&0.96/0.05&\textbf{0.99}/0.02&-0.98/0.04&\textbf{-0.99}/0.01\\
2000x20-20 10NF&-0.95/0.02&\textbf{-0.96}/0.02&0.96/0.02&\textbf{0.97}/0.01&\textbf{0.93}/0.02&\textbf{0.93}/0.01&\textbf{-0.97}/0.01&\textbf{-0.97}/0.01\\
2000x20-20 20NF&-0.93/0.02&\textbf{-0.95}/0.02&0.93/0.03&\textbf{0.96}/0.01&\textbf{0.93}/0.02&0.92/0.01&-0.96/0.01&\textbf{-0.97}/0.01\\
%2000x20-20 80NF&\textbf{-0.93}/0.02&-0.32/0.20&0.91/0.03&\textbf{0.95}/0.01&0.93/0.02&\textbf{0.94/0.01}&-0.88/0.03&\textbf{-0.94}/0.02\\
2000x30-50 15NF&-0.95/0.01&\textbf{-0.96}/0.01&0.96/0.02&\textbf{0.97}/0.01&\textbf{0.94}/0.02&\textbf{0.94}/0.01&-0.95/0.01&\textbf{-0.97}/0.01\\
2000x30-50 30NF&-0.94/0.01&\textbf{-0.95}/0.01&0.94/0.01&\textbf{0.96}/0.01&\textbf{0.93}/0.01&\textbf{0.93}/0.01&-0.94/0.01&\textbf{-0.96}/0.01\\
%2000x30-50 120NF&\textbf{-0.95}/0.01&\textbf{-0.95}/0.02&0.94/0.01&\textbf{0.96}/0.01&\textbf{0.95}/0.01&\textbf{0.95}/0.01&-0.86/0.03&\textbf{-0.92}/0.01\\
\midrule
5000x20-10 10NF&-0.97/0.01&\textbf{-0.98}/0.01&0.98/0.02&\textbf{0.99}/0.01&0.97/0.01&\textbf{0.98}/0.01&\textbf{-0.99}/0.00&\textbf{-0.99}/0.00\\
5000x20-10 20NF&-0.96/0.03&\textbf{-0.97}/0.01&0.96/0.04&\textbf{0.98}/0.01&\textbf{0.96}/0.03&\textbf{0.96}/0.01&\textbf{-0.99}/0.01&\textbf{-0.99}/0.00\\
%5000x20-10 80NF&\textbf{-0.94}/0.03&-0.41/0.22&0.91/0.06&\textbf{0.96}/0.02&0.94/0.03&\textbf{0.95}/0.01&-0.97/0.01&\textbf{-0.98}/0.01\\
5000x30-30 15NF&-0.96/0.01&\textbf{-0.97}/0.01&\textbf{0.97}/0.01&\textbf{0.97}/0.01&\textbf{0.96}/0.01&0.94/0.01&\textbf{-0.97}/0.01&\textbf{-0.97}/0.01\\
5000x30-30 30NF&\textbf{-0.95}/0.01&\textbf{-0.95}/0.01&0.95/0.02&\textbf{0.97}/0.01&\textbf{0.94}/0.01&\textbf{0.94}/0.01&\textbf{-0.97}/0.01&\textbf{-0.97}/0.01\\
%5000x30-30 120NF&\textbf{-0.93}/0.01&-0.46/0.12&0.92/0.02&\textbf{0.95}/0.01&\textbf{0.93}/0.01&\textbf{0.93}/0.01&-0.93/0.01&\textbf{-0.95}/0.01\\
5000x40-50 20NF&\textbf{-0.96}/0.01&\textbf{-0.96}/0.01&0.96/0.01&\textbf{0.97}/0.01&\textbf{0.95}/0.01&\textbf{0.95}/0.01&\textbf{-0.97}/0.01&\textbf{-0.97}/0.01\\
5000x40-50 40NF&-0.94/0.01&\textbf{-0.95}/0.01&0.95/0.01&\textbf{0.96}/0.01&\textbf{0.94}/0.01&\textbf{0.94}/0.01&-0.96/0.01&\textbf{-0.97}/0.01\\
%5000x40-50 160NF&\textbf{-0.93}/0.01&-0.89/0.05&0.93/0.02&\textbf{0.94}/0.01&\textbf{0.93}/0.01&\textbf{0.93}/0.01&-0.90/0.02&\textbf{-0.93}/0.01\\
\bottomrule
\end{tabular}}
\end{center}
\label{tab:invvar_comparison}
\end{table*}

\subsection{Real-world application}
In this section, we demonstrate how to apply our method to a real-world data set. For this example, we consider the Human Activity Recognition (HAR) data set \cite{anguita2013public}, also available in OpenML \cite{OpenML2013}. This is a complex benchmark built from recordings of 30 subjects performing activities of daily living, captured with a waist-mounted smartphone equipped with an inertial sensor. The final data set contains 10,299 data points described by 561 numerical features, partitioned into six classes corresponding to distinct activities. The HAR data set provides a challenging real-world benchmark due to its high dimensionality, sensor noise, and multi-class structure, making it well suited to evaluate whether FIR improves the reliability of internal clustering indices in practice.

We began our analysis with two important preprocessing steps. First, we ensured that no feature had a constant value across all data points (i.e., all features had positive range). We then normalised the entire data set using the range normalisation in Equation~\ref{eq:range_normalisation}. After preprocessing, we ran our experiments on this data set following the same methodology as for the synthetic data. Specifically, we applied $k$-means clustering 200 times, computing WCSS, ASW, CH, and DB, along with their FIR-enhanced versions. We repeated this entire process 50 times in order to estimate the correlation of each internal index with the ARI measured against the ground-truth class labels.

\begin{figure*}[hpt!]
    \centering
    \subfigure[Positive correlation is desirable.]{\includegraphics[width=0.45\textwidth]{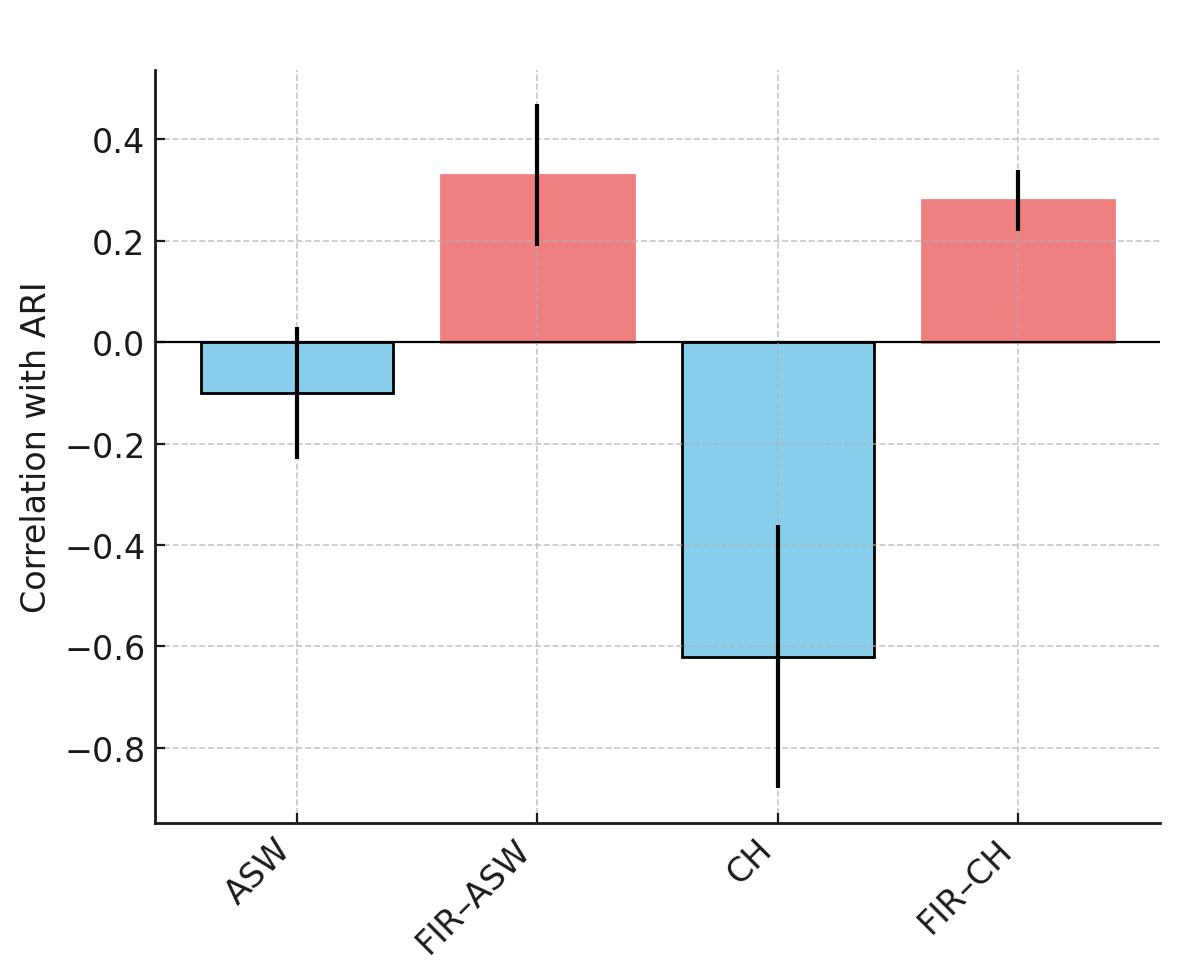}}
    \subfigure[Negative correlation is desirable.]{\includegraphics[width=0.45\textwidth]{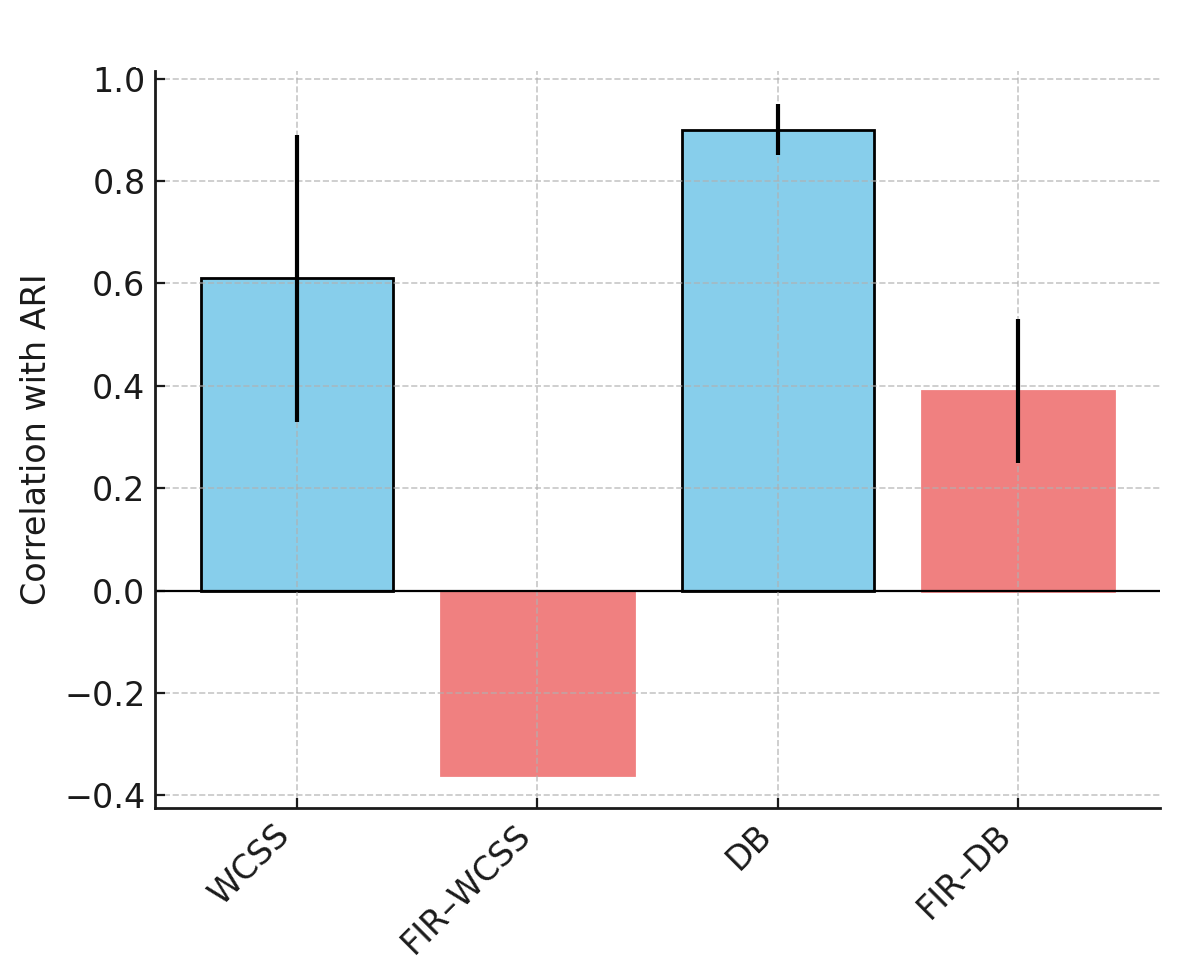}}
    \caption{Results of our experiments on the HAR data set. We executed $k$-means++ 200 times, generating a 200-component ARI vector and a corresponding 200-component vector for each index. The reported correlation measures the alignment between these index vectors and the ARI vector (i.e., ground truth). ``FIR'' represent results obtained using our proposed method. This figure shows that in all cases FIR improved the correlation.}
    \label{fig:figure_HAR}
\end{figure*}

Figure \ref{fig:figure_HAR} shows the results of our experiments. We can see that HAR is indeed a hard data set for $k$-means++ to cluster. This can be seen most clearly when noticing that we have a positive correlation between WCSS and ARI, although theoretically this correlation should be negative (i.e., low error should correlate with high accuracy). Even in such a difficult and high-dimensional data set, we can see that FIR improves the correlation.  

\section{Conclusion}

In this paper, we introduced Feature Importance Rescaling (FIR), a theoretically sound data rescaling method designed to enhance internal clustering evaluation measures by accounting for feature relevance. FIR dynamically adjusts feature scaling to better reflect each feature's contribution to cluster structure, thereby improving the reliability of commonly used internal validation indices. Through extensive experiments on synthetic data sets, we demonstrated that FIR consistently improves the correlation between internal validation measures — the $k$-means criterion, Average Silhouette Width, Calinski-Harabasz, and Davies-Bouldin — and the ground truth.

The results highlight several key findings. First, FIR is particularly beneficial in the presence of noisy or irrelevant features, significantly increasing the robustness of internal validation indices in such scenarios. Second, the improvements persist even in challenging settings where clusters exhibit a higher degree of overlap. Additionally, our results suggest that as the number of data points increases, internal validation measures become more stable, with lower variance observed across different experimental runs.

In addition to these empirical results, FIR is grounded in a clear theoretical foundation. We show that the method is strictly convex and has a unique solution for non-trivial features, that it down-weights high-dispersion features in a stable and principled way, and that it is robust to both noisy features and uniform feature rescaling. Although FIR does not satisfy the richness axiom, this is an intentional trade-off that promotes more meaningful clusterings by prioritising compactness over arbitrary flexibility.

Overall, FIR strengthens the effectiveness of internal clustering validation, offering a practical solution for real-world applications where ground truth labels are unavailable. Future work may explore its generalisability to other clustering paradigms, such as hierarchical or density-based methods, and investigate its applicability to data sets with complex feature interactions.

\bibliographystyle{ieeetr}
\bibliography{references}

\end{document}